\documentclass[lettersize,journal]{IEEEtran}  

\usepackage{amsmath} 
\usepackage{amssymb}  
\usepackage{comment}

\usepackage{sansmath}

\usepackage[dvipsnames]{xcolor}
\definecolor{mycolor0}{rgb}{1,0.62352,0.70980}
\definecolor{mycolor1}{rgb}{0.67843,0.86666,0.55686}
\definecolor{mycolor2}{rgb}{0.73725,0.74176,0.86274}
\definecolor{mycolor3}{rgb}{0.4940,0.1840,0.5560}
\definecolor{mycolor4}{rgb}{0.4660,0.6740,0.1880}
\definecolor{mycolor5}{rgb}{0.3010,0.7450,0.9330}
\definecolor{mycolor6}{rgb}{0.6350,0.0780,0.1840}
\usepackage{colortbl}
\usepackage{graphicx}
\usepackage{tikz}
\usetikzlibrary{spy}
\usetikzlibrary{patterns}
\usepackage{float}
\usepackage{makecell}
\usepackage{subcaption}
\usepackage[binary-units,detect-all]{siunitx}
\usepackage{booktabs}
\usepackage{multirow}
\usepackage{threeparttable}
\usepackage{pgfplotstable}
\usepackage{pgfplots}
\usepackage{balance} 
\usepackage{bchart}
\usepgfplotslibrary{units}
\usepackage[%
breaklinks,%
linktocpage=true,%
pdfauthor={Long Wen, Markus Rickert, Fengjunjie Pan, Jianjie Lin, Alois Knoll},%
pdfborder={0 0 0},%
pdfpagemode=UseNone,%
pdftitle={Virtualization \& Microservice Architecture for Software-Defined Vehicles: An Evaluation and Exploration}%
]{hyperref}
\usepackage{xurl}
\usepackage{array}
\usepackage{calc}
\usepackage{csquotes}
\usepackage{tabularx}
\newcolumntype{L}{>{\raggedright\arraybackslash}X}

\newcommand{\refblue}[1]{{\textcolor{blue}{\ref{#1}}}}

\definecolor{best}{rgb}{0,0.5,0.0}
\title{\LARGE\bf%
Virtualization \& Microservice Architecture for Software-Defined Vehicles: An Evaluation and Exploration%
}

\author{Long Wen$^1$, Markus Rickert$^2$, Fengjunjie Pan$^1$, Jianjie Lin$^1$, Yu Zhang$^1$, \\Tobias Betz$^3$, and Alois Knoll$^1$,~\IEEEmembership{Fellow~Member,~IEEE}
\thanks{$^1$ L. Wen, F. Pan, J. Lin, Y. Zhang, and A. Knoll are with Robotics, Artificial Intelligence and Real-Time Systems, School of Computation, Information and Technology, Technical University of Munich, Munich 85748, Germany. (email: \{wenl, panf, jianjie.lin, zha0, knoll\}@in.tum.de)}
\thanks{$^2$ M. Rickert is with Multimodal Intelligent Interaction, Faculty of Information Systems and Applied Computer Sciences, University of Bamberg, Bamberg 96050, Germany. (email: markus.rickert@uni-bamberg.de)} 
\thanks{$^3$ T. Betz is with Automotive Technology, School of Engineering and Design, Technical University of Munich, Munich 85748, Germany. (email: tobi.betz@tum.de)  
}%
}

\usetikzlibrary{external}
\usepgfplotslibrary{external}
\begin{document}

\maketitle

\begin{abstract}

The emergence of Software-Defined Vehicles (SDVs) signifies a shift from a distributed network of electronic control units (ECUs) to a centralized computing architecture within the vehicle's electrical and electronic (E/E) systems.
This transition addresses the growing complexity and demand for enhanced functionality in traditional E/E architectures, with containerization and virtualization streamlining software development and updates within the SDV framework.
While widely used in cloud computing, their performance and suitability for intelligent vehicles have yet to be thoroughly evaluated. 
In this work, we conduct a comprehensive performance evaluation of containerization and virtualization on embedded and high-performance AMD64 and ARM64 systems, focusing on CPU, memory, network, and disk metrics.
In addition, we assess their impact on real-world automotive applications using the Autoware framework and further integrate a microservice-based architecture to evaluate its start-up time and resource consumption.
Our extensive experiments reveal a slight 0-5\% performance decline in CPU, memory, and network usage for both containerization and virtualization compared to bare-metal setups, with more significant reductions in disk operations—5-15\% for containerized environments and up to 35\% for virtualized setups.
Despite these declines, experiments with actual vehicle applications demonstrate minimal impact on the Autoware framework, and in some cases, a microservice architecture integration improves start-up time by up to 18\%.   

\end{abstract}

\begin{IEEEkeywords}
	Software-defined vehicle, virtualization, microservice, benchmark.
\end{IEEEkeywords}
\section{Introduction}
\begin{figure}[t]
	\centering
	\includegraphics[width=0.48\textwidth]{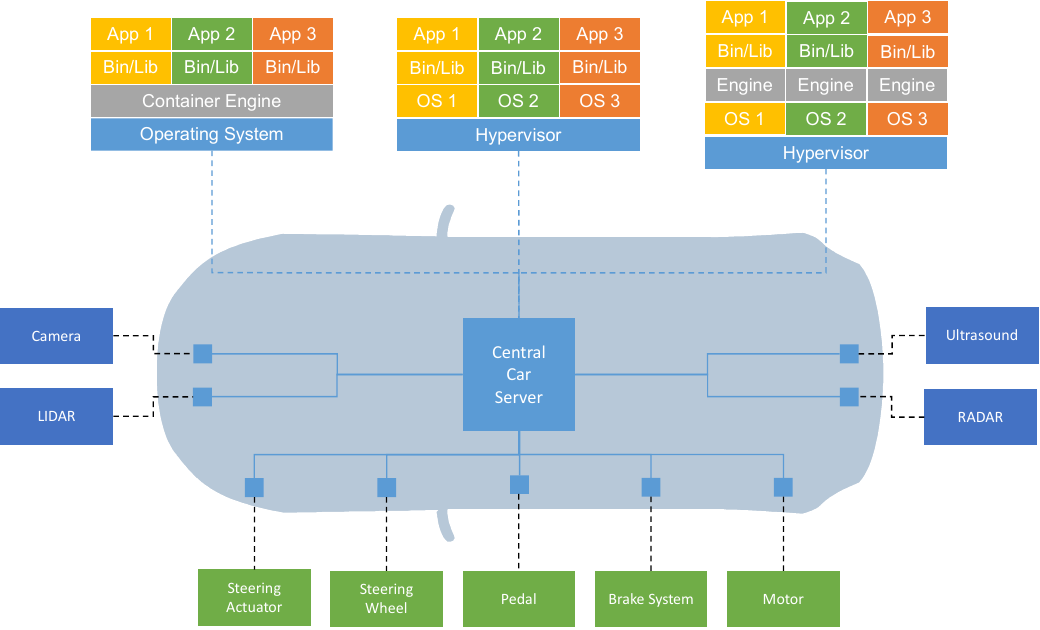}
	\caption{Potential configurations of containerization and virtualization, and their integration with microservice in SDV.}
	\label{fig:virtualdiff}
\end{figure}

\IEEEPARstart{A}{utomobiles} are becoming increasingly reliant on a large number of electronic control units (ECUs) to support enhanced functionality.
However, due to the specialized binding of software and hardware in embedded systems, modifying or updating vehicle functions in traditional electrical and electronic (E/E) architecture is complicated and time consuming~\cite{Steinkamp2021,sax2017survey}.
In addition, as intelligent vehicles continue to integrate more advanced and complex features, the traditional distributed ECU architecture will struggle to meet future demands due to its limited onboard resources~\cite{9761398,bedretchuk2023low}.
To overcome these limitations, the concept of software-defined vehicles (SDVs) has emerged, providing a flexible solution for software deployment and maintenance~\cite{Amazon,wijesekara2023optimization,abir2023digital}.
SDVs utilize scalable and adaptable architectures that decouple software from hardware~\cite{Haas2016}, enabling seamless updates and modifications to vehicle capabilities.  
Virtualization and containerization are foundational technologies in this context, essential for unlocking the flexibility and scalability of SDVs~\cite{Jain2016}.
Fig.~\refblue{fig:virtualdiff} illustrates the distinctions between virtualization and containerization within the context of SDVs, highlighting the various function applications they support. 

Virtualization and containerization are widely employed in cloud computing to manage servers, where abundant computing resources and relaxed response time requirements (ranging from 1 millisecond to 1 second) make them ideal for maintaining performance~\cite{malhotra2014virtualization}.
While these technologies can be adapted for software-defined vehicles (SDVs), the stricter resource constraints in vehicles compared to servers necessitate careful optimization~\cite{queiroz2023container}. 
For automotive applications, it is advisable to isolate security-critical software in separate virtual machines, with containerization employed within these machines to ensure efficient resource management and communication~\cite{hakak2023autonomous}.
However, despite their widespread use in the cloud domain, there is a lack of benchmark reports evaluating the performance of virtualization and containerization in automotive contexts, especially when both technologies are used simultaneously.
Additionally, as the functions in SDVs increase in complexity and number, system deployment and maintenance become more challenging.
Monolithic architectures exacerbate these difficulties by reducing system adaptability, hindering updates and modifications.


To address these challenges, this paper begins with a comprehensive performance evaluation of CPU, memory, disk, and network resources using selected benchmarking tools across various platforms, including an embedded system, a custom high-performance AMD64 workstation, and an ARM64 development platform from a real autonomous vehicle.
This evaluation provides developers with a valuable reference for selecting the most suitable virtualization and containerization technologies for optimal performance.
Furthermore, to overcome the limitations of monolithic architectures and simplify SDV software maintenance, we propose transitioning to a microservice-based architecture.
This approach decomposes the monolithic structure into smaller, independently deployable services, enabling the independent restarting and updating of specific software modules, reducing system complexity, and improving efficiency—an approach well-suited for SDVs utilizing containerization.
As illustrated in Fig.~\refblue{fig:virtualdiff}, each functional module—such as sensors, motors, and steering—can be treated as a microservice and deployed in a dedicated container.
We validate the integration of this microservice architecture within the E/E framework of SDVs by evaluating its performance on the Autoware framework in both virtualized and containerized environments on the AMD64 and ARM64 platforms. 
To thoroughly assess this approach, we implemented the microservice architecture at both module and function levels within Autoware, deploying 10, 16, and 26 dedicated containers, respectively.
Benchmark tests reveal that virtualization, containerization, and their combination with a microservice architecture deliver comparable performance to that of a bare-metal setup and can even optimize performance when utilized properly.

Our main contribution are summarized as follows:
\begin{itemize}
	\item We provide a comprehensive evaluation of containerization and virtualization technologies on both amd64 and arm64 platforms, covering generic performance metrics including CPU, memory, disk, and network as well as real-vehicle implementations.   
	\item We propose a modular solution for managing complex SDV software systems by integrating microservice architecture with virtualization. Our experimental results demonstrate the feasibility of this approach, achieving improvements up to 18\% in start-up time for the tested framework. 
\end{itemize}


\section{Related Work}

Various studies have explored the performance of virtualized and containerized environments.
Giallorenzo et al.~\cite{Giallorenzo2021} provided an in-depth review of cutting-edge hypervisors and container engines, along with a suite of reference benchmarks for high-performance computing systems. 
Xavier et al.~\cite{Xavier2013} delved into virtualization within high-performance computing, discussing the balance between performance and isolation for both containerized and virtualized setups.
Abdellatief et al.~\cite{elsayed2013performance} provided performance comparison among Type-I and Type-II hypervisors in different scenarios.
Felter et al.~\cite{felter2014ibm} focused on the hypervisor KVM and the container engine Docker for cloud computing.  
Zhanibek~\cite{KOZHIRBAYEV2017175} examined Docker and LXC's performance within the Infrastructure-as-a-Server Cloud model. 
Similarly, Raho et al.~\cite{Raho2015} assessed the system and I/O performance of KVM, Docker, and Xen on an ARM-based platform.
These studies analyzed the performance impacts of virtualization and containerization across four key metrics: CPU, memory, network throughput, and disk I/O. 
None of the previously mentioned studies have conducted benchmarks that simultaneously encompass both virtualization and containerization across different platforms.

The application of hypervisors and container engines in SDVs is a subject of growing interest.
Sundar et al.~\cite{SundarRajan2018} noted that hypervisors are capable of allocating resources to distinct virtual machines (VMs), making them suitable for mixed-criticality applications in vehicles.
This allows for the deployment of isolated VMs and containers on the same physical hardware, ensuring that various applications can operate concurrently without mutual interference.
While initiatives like SOAFEE~\cite{Adlink} are looking into the safety aspects of VMs and containers, there remains a gap in benchmark reporting on the runtime performance of these virtualization and containerization technologies in automotive scenarios.

The potential of microservice architecture to enhance the adaptability of automotive systems is explored by Benderius et al.~\cite{8057581}. 
Berger et al.~\cite{7958428} highlight both the advantages and challenges of implementing microservice architecture in software-centric sectors, such as autonomous vehicles, while also introducing their design approach. 
Lotz et al.~\cite{8712376} present an advanced driver assistance system based on microservices, illustrating its practicality and the simplification it brings to the system. 
In~\cite{9041240}, Kuhn et al. examine the architectural requirements in automotive E/E manufacturing and propose a tailored information system built on microservice architecture.
Altogether, these studies offer a theoretical perspective on the transition from traditional E/E architectures to those based on microservices, outlining the associated challenges and benefits.
However, the aforementioned mentioned research works focus solely on a certain function on the vehicle, none of them provide a practical solution for any existing automotive autonomous driving framework or platform.  





\section{Preliminaries}
While virtualization, containerization, and microservice architecture are not novel technologies, their application in the automotive industry is relatively less common. 
This section offers a concise introduction to the functions and principles of these technologies.
\subsection{Virtualization}
Virtual Machines~(VMs), the first generation of virtualization technology, date back to the 1960s.
Virtualization involves emulating hardware and software into isolated virtual computing systems known as VMs.
The software that generates and manages these virtual computers is called a hypervisor, which allocates hardware resource~(CPU, memory, and storage) and isolates execution environments across different VMs~\cite{10.1145/2644865.2541946}.
Hypervisors come in two types: Type-I, which operates directly on the hardware, and Type-II, which runs on a hosting Operating System~(OS). 
In the automotive field, Type-I hypervisors are preferred for their superior performance and enhanced security features~\cite{6871203}.
In this work, we will utilize Kernel-based Virtual Machine~(KVM), a Type-I hypervisor that enhances Linux with hypervisor capabilities. 
VMs operating under KVM are treated as standard Linux processes, allowing for seamless integration within the system's overall architecture~\cite{kivity2007kvm}.

\begin{table}[tb]
	\caption{Comparison of Docker, Podman, and Systemd-nspawn}
	\centering
	\label{table: containerdiff}
		\begin{tabular}{llll}
			\toprule
			Feature & Docker & Podman & Systemd-nspawn \\
			\midrule
			Architecture & Daemon & Daemon-less & Daemon-less \\
			Security & Root needed & Rootless & Rootless with limits \\
			Management & With Daemon & With systemd & With systemd \\
			Use Case & General & Security focus & Lightweight \\
			\bottomrule
		\end{tabular}
		\label{tab1}
\end{table}

\subsection{Containerization}
Containerization is a light-weighted virtualization that only virtualizes the OS.
A container image holds packaged, self-contained, ready-to-deploy parts of applications and, if necessary, middle-ware and business logic (in binaries and libraries) to run applications~\cite{10.1145/2723872.2723882}.
A container engine is the software that transforms container images into running containers. 
Containerized software operates uniformly across different software environments with the help of its isolation from the underlying infrastructure.
Containers share the machine’s OS kernel and therefore do not require an OS per application, driving higher efficiencies and flexibility and reducing server and licensing costs.
Docker, Podman, and systemd-nspawn are evaluated in this work, each with unique features.
Docker, the most popular, offers robust functionality and an broad ecosystem, acting as a standard in containerization~\cite{turnbull2014docker}.
Podman, similar to Docker, distinguishes itself with its daemonless architecture, enhancing security and simplicity. 
It frequently serves as a drop-in substitute for Docker.
Systemd-nspawn, less known, is coupled with Linux's systemd system and service manager.
Unlike Docker and Podman, which are broadly aimed at application containers, systemd-nspawn is more closely aligned with Linux OS environments, with a focus on system-level resource isolation.
A detailed comparison of these technology is presented in Table~\refblue{table: containerdiff}.
\subsection{Microservice Achitecture}
Microservice architecture, a predominant architectural style in service-oriented software, emphasizes dividing a system into small, cohesive services each designed to perform a specific function~\cite{Dragoni2017}.
This architecture can be viewed as a distributed application where each module is a microservice.
The widely recognized benefits include increased agility, developer productivity, resilience, scalability, reliability, maintainability, separation of concerns, and ease of deployment~\cite{7796008}.
The inherent modularity and scalability of microservices make them ideally compatible with virtualization and containerization technologies.
These technologies facilitate the deployment and management of microservices by providing isolated, flexible environments that can be dynamically allocated and scaled.
This synergy enhances the overall agility and efficiency of software systems, making the combination of microservice architecture and virtualization ideal for the complex E/E architecture of SDVs.

\section{Generic Performance Benchmark}
In this section, we present a general performance evaluation of virtualization and containerization tools regarding CPU, memory, disk, and network on different platforms.
\subsection{Experimental Setup}

\begin{table*}[tb]
	\caption{Configuration of platforms used in experiments.}
	\centering
	\label{table: config1}
	\sisetup{table-number-alignment=right,table-parse-only=true}%
	\setlength{\tabcolsep}{0.7\tabcolsep}
	\begin{tabularx}{\linewidth}{p{1.2cm}LLLLLLLp{0.8cm}}
		\toprule
		& Kernel & CPU & Cores (HT) & RAM & Disk& Disk 2&Disk 3 & PCIe \\
		\midrule
		Embedded & 5.15.0-1013-raspi & ARM Cortex-A72 & 4(4) \SI{1.5}{\giga\hertz} & \SI{4}{\giga\byte} LPDDR4 \SI{3200}{\mega\hertz} &Crucial MX500 SATA \SI{250}{\giga\byte}&-&-&- \\

		
		Workstation& 5.15.0-46-generic & Intel i9-12900K & 8(16) \SI{3.2}{\giga\hertz}\par 8(\hphantom{1}8) \SI{2.4}{\giga\hertz}& \SI{32}{\giga\byte} DDR5 \SI{5200}{\mega\hertz}  & Crucial MX500 SATA \SI{250}{\giga\byte}&Samsung 980 Pro M.2 \SI{500}{\giga\byte}&Intel D7 P5520 U.2 \SI{1.92}{\tera\byte}&4.0\\
		
		AVA& 5.15.0-46-generic & Ampere Altra Q32-17 & 32(32) \SI{1.5}{\giga\hertz}\par max.\hphantom{1.} \SI{1.7}{\giga\hertz}& \SI{32}{\giga\byte} DDR4 \SI{3200}{\mega\hertz} & Samsung 980 Pro M.2 \SI{2}{\tera\byte}&-&-&4.0\\
		
		\bottomrule
	\end{tabularx}
\end{table*}

\begin{figure}[tb]
	\centering
	\sffamily
	\sansmath
	\footnotesize
	\begin{tikzpicture}
		\newcounter{nodecount}
		\setcounter{nodecount}{0}
		\begin{axis}[
			ybar,
			width=8.5cm,
			height=5.5cm,
			ymin=0,
			ymax=100,
			ymajorgrids,
			enlarge x limits=0.2,
			nodes near coords,
			every node near coord/.append style={
				font=\tiny,/pgf/number format/.cd,fixed zerofill,precision=1,
				/tikz/execute at begin node={%
					\stepcounter{nodecount}
					\ifnum\thenodecount=1 \color{blue} \fi 
					\ifnum\thenodecount=2 \color{blue} \fi 
					\ifnum\thenodecount=3 \color{blue} \fi 
				}
			},
			axis line style={draw=none},
			every axis plot/.append style={fill,draw=none,no markers},
			ylabel=Percentage compared to bare-metal,
			legend image code/.code={\draw [#1,draw=none] (0cm,-0.5em) rectangle (0.5em,0.5em);},
			legend style={font=\scriptsize, legend columns=-1, at={(0.5,1.2)},anchor=north, draw=none, fill=none},
			symbolic x coords={Bare-metal, KVM-qcow2, KVM-raw, KVM-partition},
			xtick=data,
			xtick style={draw=none},
			ytick style={draw=none},
			]
			\addplot[fill=mycolor0] coordinates {
				(KVM-qcow2,   65.02)
				(KVM-raw,  66.04)
				(KVM-partition,   68.76)
			};
			\addplot[fill=mycolor1] coordinates {
				(KVM-qcow2,   59.37)
				(KVM-raw,  62.59)
				(KVM-partition,   64.65)
			};
			\addplot[fill=mycolor2] coordinates {
				(KVM-qcow2,   60.22)
				(KVM-raw,  61.54)
				(KVM-partition,   63.10)
			};
			\legend{Embedded, AVA, Workstation}
			
		\end{axis}
	\end{tikzpicture}
	\caption{Performance comparison between disk image formats.}
	\label{fig:diskimage}
\end{figure}
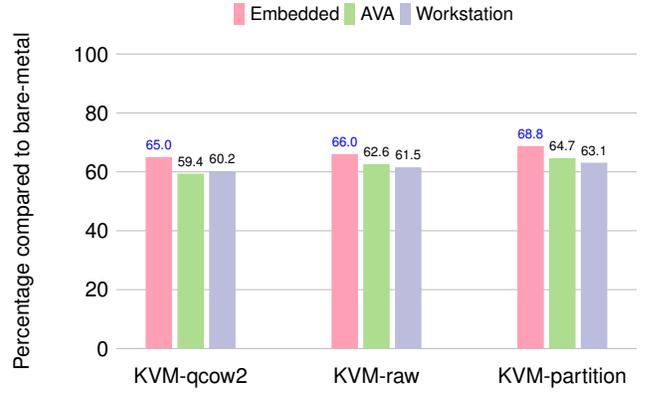

To assess CPU performance, we employ benchmarking tools such as Whetstone~(v1.2)~\cite{Curnow1976}, Dhrystone~(v2.2a)~\cite{Weicker1984}, and Kcbench~(v0.9.5)~\cite{Leemhuis2022}.
RAMspeed~(v3.5.0) is utilized for memory performance testing, while network throughput is obtained using iPerf3~(v3.9). 
Disk I/O is evaluated with the aid of Dbench~(v4.00) and Bonnie++~(v2.00). 
The specific functions of each benchmark tool will be detailed in the subsequent section. 
Our evaluation encompasses a range of hypervisors and container engines, including Docker~(v20.10.17), KVM~(v6.2.0), Podman~(v3.4.4), and Systemd-Nspawn~(v249.11).

The experiments are performed on three platforms with different hardware configurations: Embedded~(Raspberry Pi 4 Model B), Workstation~(a high-performance custom AMD64 workstation), and AVA~(an ARM64 developer platform on a real autonomous vehicle provided by SOAFEE). 
Detailed specifications of the platform configurations and software versions are provided in Table~\refblue{table: config1}.
For the bus interface benchmark test on the Workstation, we used a variety of disks, listed as Disk 1-4 in Table~\refblue{table: config1}.
Different disk types were selected to optimize the performance of each platform. 
Our results are reported as percentage differences relative to a bare-metal reference standard.

The configuration of the KVM hypervisor is standardized across all platforms.
On each, we establish a single virtual machine, dedicating all available CPU cores, GPU and RAM to it.
The Ubuntu system on the virtual machine is configured in server mode to minimize resource usage by disabling the graphical user interface. 
For disk virtualization, we opt for the \emph{native} I/O mode and configure the cache setting to \emph{none}.
Additionally, we observe that disk image format impacts disk performance obviously.
We compare three disk image formats: raw, qcow2, and using a partition directly as a disk image.
Our preliminary experiment, depicted in Fig.~\refblue{fig:diskimage}, indicates that using a partition as a disk image provides superior performance on all tested platforms.  
Therefore, we employ the partition disk format in all the following experiments.
During the evaluation of containerization technology, the default settings of container engines are utilized. 

\begin{table}[tb]
	\caption{Network performance of KVM on three platforms. Numbers specify percentages compared to the bare-metal reference~(\SI{100}{\percent}).}
	\centering
	\label{table: Net}
	\sisetup{table-number-alignment=right,table-parse-only=true}%
	\begin{tabular}{lSSS}
		\toprule
		& {Embedded} & {AVA} & {Workstation} \\
		\midrule
		iPerf3 send & 96.67 & \textbf{97.31}  & {97.13}\\
		iPerf3 receive & 94.47 & 99.62 & \textbf{99.68} \\
		\bottomrule
	\end{tabular}
	\vspace{-1em}
\end{table}

\subsection{CPU Performance}
\subsubsection{Benchmark tools} 
Whetstone serves to evaluate the floating-point operation performance of CPUs.
We run Whetstone 50 million times to ensure meaningful and stable results.
Dhrystone, a synthetic computing benchmark, focuses on the CPU's integer operations by reporting the number of completed main code loop iterations per second.
Kcbench, a script for benchmarking, measures the time taken to compile the Linux kernel multiple times, providing a reasonable assessment of the CPU's comprehensive performance.
In our tests, we compile the Linux 4.11 kernel source code. 
While the default job count is twice the number of CPU cores, it requires fine-tuning for different CPU types.
For our evaluation, we set the job count to 4 for the quad-core CPUs of the embedded platform and to 24 and 32 for the amd64 and arm64 Workstations.

\subsubsection{Benchmark results}
\begin{figure*}[tb]
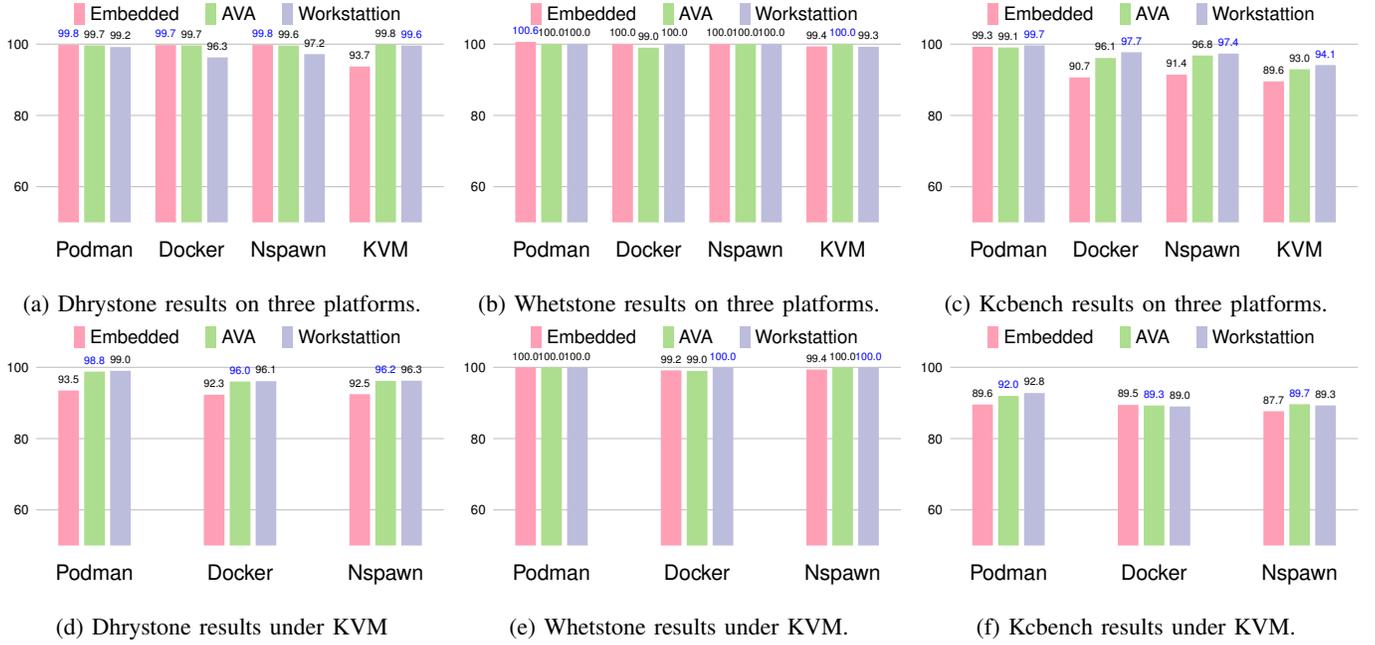

	\centering
	\sffamily
	\sansmath
	\footnotesize
	\begin{subfigure}{0.33\textwidth}
		\centering
		 

		\caption{Dhrystone results on three platforms.}
		\label{fig:CPU-Dhrystone}
	\end{subfigure}%
	\hfill%
	\begin{subfigure}{0.33\textwidth}%
		\centering
		 
		%
	\caption{Whetstone results on three platforms.}
	\label{fig:CPU-Whetstone}
	\end{subfigure}%
	\hfill%
	\begin{subfigure}{0.33\textwidth}%
		\centering
		 
		%
		\caption{Kcbench results on three platforms.}
		\label{fig:CPU-Kcbench}
	\end{subfigure}%
	\hfill%
	\begin{subfigure}{0.33\textwidth}%
		\centering
		
		%
		\caption{Dhrystone results under KVM}
		\label{fig:CPU-Dhrystone-kvm}
	\end{subfigure}%
	\hfill%
	\begin{subfigure}{0.33\textwidth}%
		\centering
		%
		\caption{Whetstone results under KVM.}
		\label{fig:CPU-Whetstone-kvm}
	\end{subfigure}%
	\hfill%
	\begin{subfigure}{0.33\textwidth}%
		\centering
		%
		\caption{Kcbench results under KVM.}
		\label{fig:CPU-Kcbench-kvm}
	\end{subfigure}%
	\caption{CPU performance of four technologies. Numbers specify percentages compared to the bare-metal reference~(\SI{100}{\percent}).}
	\label{fig:CPU}
\end{figure*}

The results are documented in Fig.~\refblue{fig:CPU}, where the values represent percentages relative to the respective platform's bare-metal reference.
It is important to note that comparable percentages across different platforms do not imply equivalent absolute performance levels.
The CPU performance gap between the three platforms is significant, as is the memory and disk I/O performance gap. 
These four virtualization and containerization technologies perform nearly identically in terms of float-point and integer performance compared to bare metal, with the Workstation delivering the least performance loss.
We collect the performance of containerization technologies in a virtualized environment in Figs.~\refblue{fig:CPU-Dhrystone-kvm}-\refblue{fig:CPU-Kcbench-kvm}.
These experiments model the scenario where applications are isolated from each other with the help of VM to meet specific requirements, such as ensuring environmental safety.
The findings reveal a noticeable performance decline when employing both virtualization and containerization layers, with Kcbench recording the most significant drop at 10\%.
  
One interesting finding is that Podman outperforms bare-metal configurations in the Whetstone benchmark when tested on Embedded systems.
This superior performance may be attributed to the different kernel configurations utilized in the container and bare-metal tests. 
Notably, there is no dedicated \emph{Ubuntu for Raspberry Pi} container image available, leading to the use of a generic Ubuntu image for container tests.
The Kcbench benchmarks indicate that KVM exhibits the most significant performance degradation compared to other technologies across all three tested platforms.
Notably, the Embedded platform shows a higher degree of performance loss relative to the other platforms.
In contrast, Podman's performance loss remains consistently moderate throughout the CPU-intensive tests.

\begin{figure*}[tb]
	\centering
	\sffamily
	\sansmath
	\footnotesize
	\vspace{-0.2cm}
	\begin{subfigure}{0.33\textwidth}
		\centering
		%
		\caption{Memory int results on three platforms.}
		\label{fig:Memory-int}
	\end{subfigure}%
	\hfill%
	\begin{subfigure}{0.33\textwidth}%
		\centering
		%
		\caption{Memory float results on three platforms.}
		\label{fig:Memory-float}
	\end{subfigure}%
	\hfill%
	\begin{subfigure}{0.33\textwidth}%
		\centering
		%
		\caption{Memory mixed results on three platforms.}
		\label{fig:Memory-mix}
	\end{subfigure}%
	\hfill%
	\begin{subfigure}{0.33\textwidth}%
		\centering
		%
		\caption{Memory int results under KVM.}
		\label{fig:Memory-int-KVM}
	\end{subfigure}%
	\hfill%
	\begin{subfigure}{0.33\textwidth}%
		\centering
		%
		\caption{Memory float results under KVM.}
		\label{fig:Memory-float-KVM}
	\end{subfigure}%
	\hfill%
	\begin{subfigure}{0.33\textwidth}%
		\centering
		%
		\caption{Memory mixed results under KVM.}
		\label{fig:Memory-mix-kvm}
	\end{subfigure}%
	\caption{Memory performance of four technologies. Numbers specify percentages compared to the bare-metal reference~(\SI{100}{\percent}).}
	\label{fig:Memory}
	\vspace{-0.5em}
\end{figure*}

\begin{figure*}[tb]
	\centering
	\sffamily
	\sansmath
	\footnotesize
	\vspace{-0.2cm}
	\begin{subfigure}{0.33\textwidth}
		\centering 
		%
		\caption{Disk I/O results on three platforms.}
		\label{fig:Disk-i/o}
	\end{subfigure}%
	\hfill%
	\begin{subfigure}{0.33\textwidth}%
		\centering
		 
		%
		\caption{Disk create results on three platforms.}
		\label{fig:Disk-create}
	\end{subfigure}%
	\hfill%
	\begin{subfigure}{0.33\textwidth}%
		\centering
		%
		\caption{Disk delete results on three platforms.}
		\label{fig:Disk-delete}
	\end{subfigure}%
		\hfill%
	\begin{subfigure}{0.33\textwidth}%
		\centering
		%
		\caption{Disk I/O results under KVM.}
		\label{fig:Disk-I/O-KVM}
	\end{subfigure}%
	\hfill%
	\begin{subfigure}{0.33\textwidth}%
		\centering
		
		%
		\caption{Disk create results under KVM.}
		\label{fig:Disk-create-KVM}
	\end{subfigure}%
	\hfill%
	\begin{subfigure}{0.33\textwidth}%
		\centering
		%
		\caption{Disk delete results under KVM.}
		\label{fig:Disk-delete-kvm}
	\end{subfigure}%
	\caption{Disk performance of four technologies. Numbers specify percentages compared to the bare-metal reference~(\SI{100}{\percent}).}
	\label{fig:Disk}
\end{figure*}

\subsection{Memory Performance}
\subsubsection{Benchmark tool} 
RAMspeed measures a computer system's cache and memory performance. 
We use version~3.5.0 for multiprocessor machines running UNIX-like operating systems~\cite{Hollander2002}.
We collect read and write speeds for integer and floating-point operations, in addition to Copy, Scale, Add, and Triad operations. 
Each measurement is repeated ten times per run, and the final result is derived from the average of five such runs.
\subsubsection{Benchmark results}
In Fig.~\refblue{fig:Memory}, int and float results represent the performance metrics for integer and floating-point read-write operations, respectively, while the mixed results reflect the performance of Copy, Scale, Add, and Triad operations. 
The data indicates that containerization incurs minimal memory performance loss, while KVM exhibits a bigger reduction in performance compared to container engines. 
Specifically, on the Embedded platform, KVM shows over a 10\% decline in memory performance relative to a bare-metal setup.
Therefore, the performance of container engines under KVM is around 10\% lower than the bare-metal. 

\subsection{Network Performance}
\subsubsection{Benchmark tool} 
iPerf3 is a tool for active measurements of the maximum achievable bandwidth on IP networks~\cite{Dugan}.
Both TCP send and receive performances are assessed, with each test running for 180 seconds.
To ensure reliable outcomes, we repeat each test 10 times.
The bit-rate limit is removed to avoid any constraints on the measurements.
Furthermore, we omit the initial 10 seconds of each test to avoid the TCP slow-start phase, allowing for an assessment of peak performance.
\subsubsection{Benchmark result} 
Table~\refblue{table: Net} exclusively reports the results for KVM, as the container engines consistently deliver stable performance~(100\%) across the three platforms. 
The KVM slightly influences the network performance on these platforms.
Furthermore, deploying containerization on top of KVM shows network performance identical to using KVM alone.

\subsection{Disk Performance}
\subsubsection{Benchmark tools}
Dbench is a benchmarking tool that generates workloads for file systems to measure the throughput~\cite{Tridgell}.
It operates using a load file that specifies a sequence of file operations, such as open, read, and close. 
We employed the default load file for our experiments, which executes a series of operations, including file creation, writing, and saving.
Like Kcbench, the process count affects Dbench's results.
The number of processes is set to 4 for the Embedded platform, 24 for the Workstation platform, and 32 for the AVA platform. 
Moreover, we utilize Sysbench, a benchmarking tool that can perform benchmarks with the \emph{fsync} function both enabled and disabled. 
We also conduct evaluations using Bonnie++~\cite{Coker2001}, which specifically tests the performance of file I/O operations and file creation or deletion.
The number of files for creation and deletion is set to 512, and a consistent random seed is set to ensure that Bonnie++ executes the tests uniformly in each run. 

\subsubsection{Benchmark results}
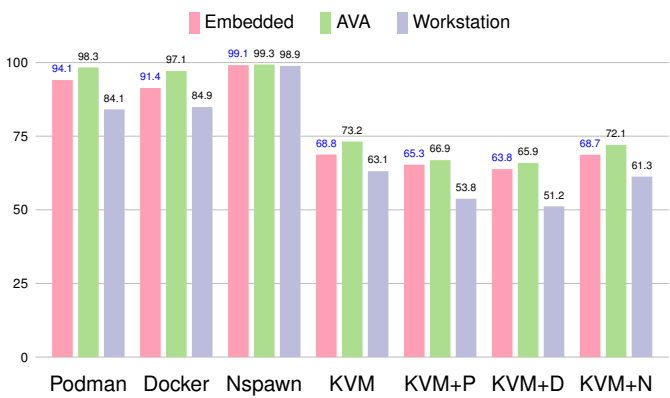
\begin{figure}[tb]
	\centering
	\sffamily
	\sansmath
	\footnotesize
	\begin{tikzpicture}
		\newcounter{nodecount}
		\setcounter{nodecount}{0}
		\begin{axis}[
			ybar,
			width=10cm,
			height=5.5cm,
			bar width=7.8pt,
			ymin=0,
			ymax=100,
			ymajorgrids,
			nodes near coords,
			every node near coord/.append style={
				font=\fontsize{4}{5}\selectfont,/pgf/number format/.cd,fixed zerofill,precision=1,
				/tikz/execute at begin node={%
					\stepcounter{nodecount}
					\ifnum\thenodecount=1 \color{blue} \fi 
					\ifnum\thenodecount=2 \color{blue} \fi
					\ifnum\thenodecount=3 \color{blue} \fi
					\ifnum\thenodecount=4 \color{blue} \fi 
					\ifnum\thenodecount=5 \color{blue} \fi
					\ifnum\thenodecount=6 \color{blue} \fi
					\ifnum\thenodecount=7 \color{blue} \fi 
				}
			},
			axis line style={draw=none},
			every axis plot/.append style={fill,draw=none,no markers},
			cycle list={mycolor0,mycolor1,mycolor2,mycolor3,mycolor4,mycolor5,mycolor6},
			ylabel={},
			ylabel style={yshift=-6em},
			legend image code/.code={\draw [#1,draw=none] (0cm,-0.5em) rectangle (0.5em,0.5em);},
			legend style={font=\scriptsize,legend columns=-1, at={(0.5,1.2)},anchor=north, draw=none, fill=none,/tikz/every even column/.append style={column sep=8pt}},
			symbolic x coords={Podman, Docker, Nspawn, KVM, KVM+P, KVM+D, KVM+N},
			xtick=data,
			xtick style={draw=none},
			ytick={0,25,50,75,100},
			yticklabel style={font=\tiny},
			ytick style={draw=none},
			]
			\addplot[fill=mycolor0] coordinates {
				(Podman,   94.06)
				(Docker,  91.35)
				(Nspawn,   99.12)
				(KVM,  68.77)
				(KVM+P,  65.32)
				(KVM+D,  63.82)
				(KVM+N,  68.69)
			};
			\addplot[fill=mycolor1] coordinates {
				(Podman,   98.33)
				(Docker,  97.12)
				(Nspawn,   99.33)
				(KVM,  73.18)
				(KVM+P,  66.87)
				(KVM+D,  65.92)
				(KVM+N,  72.05)
			};
			\addplot[fill=mycolor2] coordinates {
				(Podman,   84.10)
				(Docker,  84.88)
				(Nspawn,   98.86)
				(KVM,  63.10)
				(KVM+P,  53.77)
				(KVM+D,  51.17)
				(KVM+N,  61.27)
			};
			\legend{Embedded, AVA, Workstation}
		\end{axis}
	\end{tikzpicture}
	\caption{Dbench results of four technologies on different platforms. Numbers specify percentages compared to the bare-metal reference~(\SI{100}{\percent}).}
	\label{fig:disk}
\end{figure}

The Dbench benchmarks, as displayed in Table Table~\refblue{fig:disk}, indicate that Nspawn outperforms the other technologies, operating at 99\% efficiency compared to bare-metal. 
KVM experiences the most significant performance drop of 35\%, while Podman and Docker exhibit moderate losses, ranging between 6\% and 15\%.
Notably, the performance further increases to over 40\% loss when KVM is combined with container technologies.

\begin{figure}[tb]
	\centering
	\sffamily\sansmath
	\footnotesize
	\begin{tikzpicture}
		\begin{axis}[
			ybar,
			ymin=0,
			ymax=6000,
			width=9cm,
			height=5.5cm,
			ymajorgrids,
			bar width= 9pt,
			ybar=3pt,
			ylabel={},
			ylabel style={yshift=-6em},
			every axis y label/.style={at={(ticklabel* cs:1.0)}},
			nodes near coords,
			nodes near coords style={font=\tiny,/pgf/number format/.cd,fixed zerofill,precision=1,/pgf/number format/1000 sep=},
			axis line style={draw=none},
			every axis plot/.append style={fill,draw=none,no markers},
			cycle list={mycolor0,mycolor1,mycolor2,mycolor3,mycolor4,mycolor5,mycolor6},
			legend cell align={left},
			legend image code/.code={\draw [#1,draw=none] (0cm,-0.5em) rectangle (0.5em,0.5em);},
			legend style={font=\scriptsize, legend columns=1, at={(0.175,0.95)},anchor=north, draw=none, fill=none},
			symbolic x coords={CR-SATA, SA-PCIe 4.0, WD-PCIe 4.0, IN-PCIe 4.0},
			y tick label style={/pgf/number format/1000 sep=},
			xtick=data,
			ytick={0,1000,2000,3000,4000,5000,6000},
			xtick style={draw=none},
			ytick style={draw=none},
			enlarge x limits=0.2,
			]
			\addplot[fill=mycolor3] coordinates {
				(CR-SATA,   42.45)
				(SA-PCIe 4.0,  19.69)
				(WD-PCIe 4.0, 91.76)
				(IN-PCIe 4.0,   2777.23)
			};
		    \addplot[fill=mycolor4] coordinates {
		    	(CR-SATA,   28.29)
		    	(SA-PCIe 4.0,  13.13)
		    	(WD-PCIe 4.0, 61.18)
		    	(IN-PCIe 4.0,   1851.49)
		    };
			\addplot[fill=mycolor5] coordinates {
				(CR-SATA,   245.01)
			    (SA-PCIe 4.0,  5029.27)
		    	(WD-PCIe 4.0, 5325.69)
		    	(IN-PCIe 4.0,   4279.72)
			};
		    \addplot[fill=mycolor6] coordinates {
		    	(CR-SATA,   163.33)
		    	(SA-PCIe 4.0,  3352.86)
		    	(WD-PCIe 4.0, 3550.46)
		    	(IN-PCIe 4.0,   2853.17)
		    };
			\legend{Read w/ fsync, Write w/ fsync, Read w/o fsync, Write w/o fsync}
			
		\end{axis}
	\end{tikzpicture}
	\caption{Performance comparison between disks (\SI{}{\mega\byte}/\SI{}{\second})}
	\label{fig:diskbus}
\end{figure}
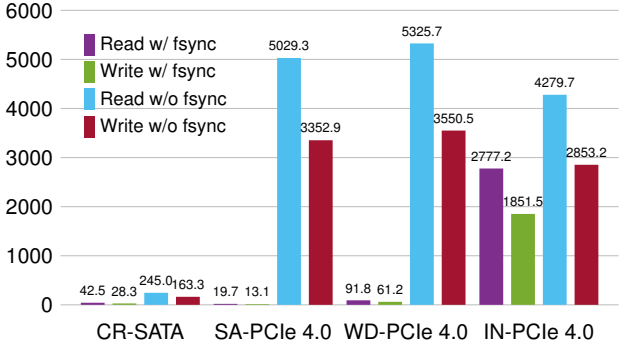

In addition, we explore how synchronization~(\emph{fsync}) affects disk performance.
The four types of disks tested are detailed in Table~\refblue{table: config1}, and the outcomes are illustrated in Fig.~\refblue{fig:diskbus}.
For these experiments, we employed Sysbench~(v1.0.20) to obtain results that are easily interpretable.
The results indicate that using \emph{fsync} markedly reduces the performance of the first three disks. 
Applications employ \emph{fsync} to guarantee that all data is written from the buffer to the disk up until the moment the \emph{fsync} is invoked.
This precaution ensures that, in the event of a system crash, the data remains intact.
Certain disks, like the enterprise-grade Intel SSD tested in our experiment, are optimized for \emph{fsync}, resulting in superior performance when \emph{fsync} is enabled.
The performance of the Intel disk without \emph{fsync} is 15\% lower compared to the Samsung and Western Digital disks, suggesting that the \emph{fsync} optimization negatively impacts disk performance.
Consequently, whether the chosen application applies \emph{fsync} will affect the disk selection.

Apart from fysnc, we evaluated the impact of the bus interface on performance. 
The data presented in Fig.~\refblue{fig:disk} are collected on Embedded~(SATA), AVA~(PCIe 4.0), and Workstation~(PCIe 4.0) with different bus interfaces. 
The performance loss on the Desktop and Workstation platforms is almost identical.
Interestingly, the Embedded platform shows even less performance loss through the SATA interface despite its inherently slower speed, with the exception of the Nspawn test.
This is likely because its benchmark performance is constrained by other factors, such as CPU performance, resulting in a lower loss.
In conclusion, the bus interface has a minimal impact on virtualization and containerization performance.  
Thus, the choice of bus interface and disk type can be tailored to meet specific user needs.

We describe the results of Bonnie++ in Fig.~\refblue{fig:Disk}. 
Nspawn exhibits the smallest performance loss compared to other technologies. 
KVM performs well in I/O and creation tasks, with a modest performance loss of 5-10\%, yet it suffers a significant decline during deletion operations across all three platforms.
Docker and Podman, on the other hand, show the poorest performance in the Bonnie++ benchmark, particularly in the creation task, where they experience a 40\% loss.
When combining KVM with containerization, all tested technologies show unsatisfying performance in creating and deleting tasks.
Therefore, it is advisable to avoid running applications with high file system churn on this setup.

\section{Automotive Scenario Benchmark}

\begin{table}[tb]
	\caption{Details of Autoware modules.}
	\centering
	\label{table: module}
	\sisetup{table-number-alignment=right,table-parse-only=true}%
	\begin{tabular}{llr}
		\toprule
		Module& Functions & Nodes \\
		\midrule
		Vehicle & robot\_state\_publisher & 1 \\
		System & system\_monitor & 12 \\
		Sensing & \makecell[tl]{Lidar 1-4, IMU\_driver, GNSS\_driver, \\Velocity\_converter } & 32 \\
		Planning & \makecell[tl]{Mission\_planning, Scenario\_selector,\\Velocity\_planning, Scenario\_library } & 26 \\
		Perception & \makecell[tl]{Obstacle\_segmentation, obstacle\_recognition, \\Traffic\_light\_recognition, occupancy\_grid\_map } & 36 \\
		Map & Map\_loader & 7 \\
		Location & \makecell[tl]{Pose\_twist\_estimator, pose\_twist\_fusion\_filter } & 12 \\
		Control & \makecell[tl]{Velocity\_controller, Vehicle\_cmd\_gate\\shift\_decider, latlon\_muxer} & 8 \\
		Api & \makecell[tl]{AD\_API, API\_adapter, API\_utils, Web\_controller} & 62 \\
		Rviz & Visualization & 1 \\
		\bottomrule
	\end{tabular}
\end{table}

\begin{figure}[t]
	\centering
	\includegraphics[width=0.48\textwidth]{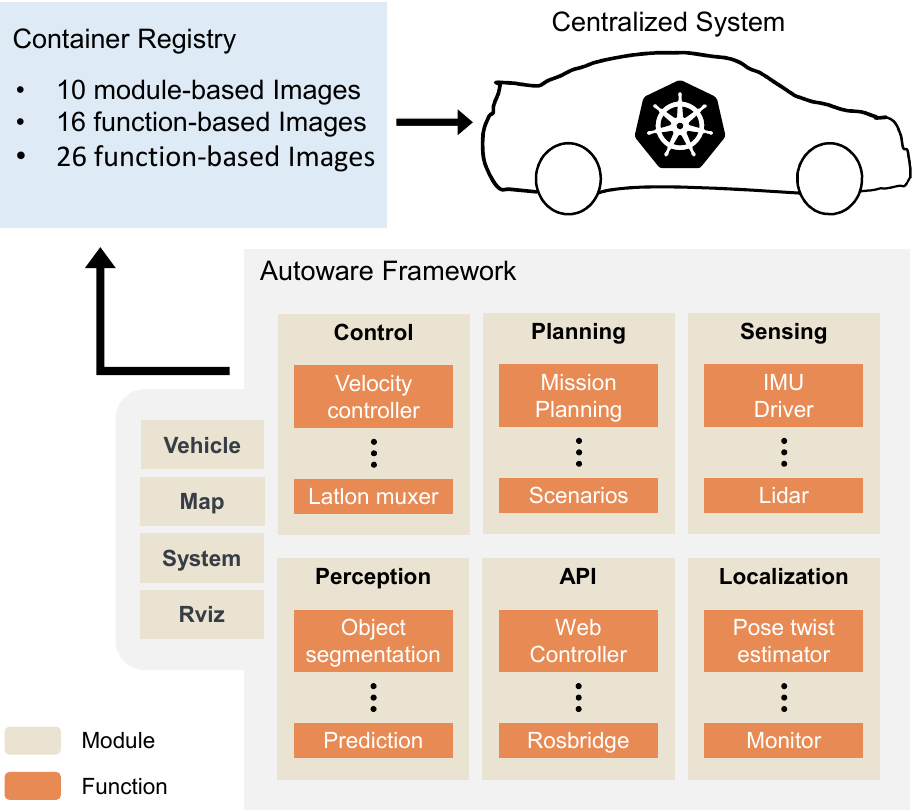}
	\caption{Microservice architecture for Autoware framework}
	\label{fig:microservice}
\end{figure}

The performance evaluation from the previous section indicates that both ARM64 and AMD64-based systems using hypervisors and container engines offer performance comparable to bare-metal configurations.
This section expands our research to include a practical automotive context by employing the Autoware framework.
We measure performance based on application start-up time, CPU, memory usage.
Additionally, we introduce a enhanced microservice architecture for the Autoware framework by leveraging containerization.
We decompose the framework into modular components and deploy each within a dedicated container to assess the performance impact of this architecture.
Experiments are also carried out in hybrid environments that utilize both virtual machines and containers.     

\subsection{Experimental Setup}
Autoware is an advanced autonomous driving framework built on ROS~(Galactic), comprising 242 ROS nodes that encompass all essential functionalities for operating an autonomous vehicle. 
The inter-device communication facilitated by ROS enables the segmentation of Autoware into various containers.
To get a comprehensive result, we establish two microservice architectures that divide the framework according to module and function, referred to as the modular level and functional level architectures, respectively.
The modular level architecture consists of nine Autoware modules to simulate the automotive scenario, including vehicle, map, sensing, perception, location, planning, control, api, and rviz.
Each module is deployed into a separate container.
Additionally, we set up a rosbag container responsible for playing back the point-cloud data recorded in a realistic scenario.
Adopting the microservice architecture, we have deconstructed the functionalities of each module into discrete parts, forming two function-level architectures with a total of 17 and 26 components, respectively.
The architecture with 17 containers specifically divides the perception and planning modules, which encompass the most content to launch. 
The 26-container architecture further splits all modules according to their individual functions.
Table~\refblue{table: module} provides detailed information and outlines the functions in these modules. 

Additionally, Fig.~\refblue{fig:microservice} presents the comprehensive architecture implemented in the benchmakring. 
The Workstation and AVA platforms are selected to provide an overview of the performance of both x86-64 and ARM architectures and we will refer these two platforms as \emph{amd64} and \emph{arm64} for clarity.

\subsubsection{Choice of Disk} 
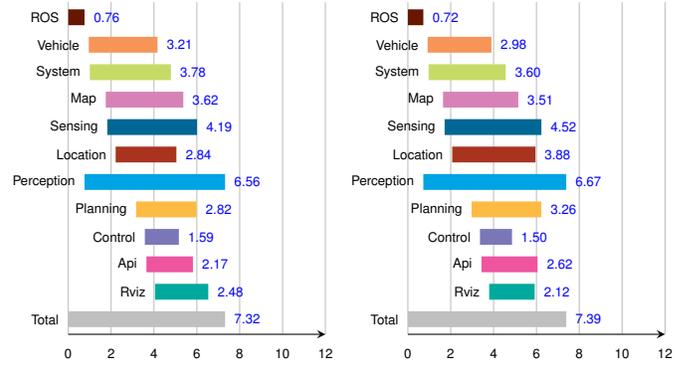
\begin{figure}[t]
	\centering
	\pgfplotsset{
		enlarge y limits=0.05,
		every axis/.append style={font=\sansmath\sffamily\tiny},
		xlabel={},
		every axis x label/.style={at={(ticklabel* cs:1.0)},anchor=south east},
		height=6cm,
		width=5cm,
		xmax=12,
		xtick={0,2,4,6,8,10,12},
		ytick={0,1,2,3,4,5,6,7,8,9,10,11,12},
	}
	\begin{subfigure}{0.24\textwidth}%
		\centering
		\begin{tikzpicture}
			\pgfplotstableread{
				i  Module     Start  Duration Color     	 
				0  ROS        0      0.759	  Sepia	
				1  Vehicle    0.956  3.209	  Peach	
				2  System     1.005  3.784	  SpringGreen
				3  Map        1.749  3.619	  Thistle	
				4  Sensing    1.823  4.187	  MidnightBlue	
				5  Location   2.211  2.835	  Mahogany	
				6  Perception 0.759  6.559    Cerulean	
				7  Planning   3.167  2.818    Dandelion	
				8  Control    3.573  1.588	  Periwinkle
				9  Api        3.648  2.169	  Rhodamine
				10 Rviz       4.050  2.484    Emerald
				11 Total      0      7.318    lightgray 
			}\datatable
			\pgfplotstablegetrowsof{\datatable}
			\pgfmathsetmacro{\tablerows}{int(\pgfplotsretval-1)}
			\begin{axis}[
				xbar stacked,
				axis x line=bottom,
				axis y line=none,
				bar width=6pt,
				every axis plot/.append style={fill,draw=none,no markers},
				nodes near coords,
				nodes near coords align={right},
				nodes near coords style={/pgf/number format/.cd,fixed zerofill,precision=2},
				point meta=explicit,
				xmajorgrids,
				xtick style={draw=none},
				ymax=11,
				ymin=0,
				y dir=reverse,
				yticklabels from table={\datatable}{Module},
				ytick style={draw=none},
				minimum/.style={forget plot,draw=none,fill=none},
				select row/.style={
					x filter/.code={\ifnum\coordindex=#1\else\def\pgfmathresult{}\fi}
				},
				]
				\addplot [minimum,nodes near coords align={left},point meta=explicit symbolic] table [x=Start,y=i,meta=Module] {\datatable};
				\addplot +[draw=none,fill=none] table [x=Duration,y=i,meta=Duration] {\datatable};
				\pgfplotsinvokeforeach{0,...,\tablerows}{
					\pgfplotstablegetelem{#1}{Color}\of{\datatable}
					\expandafter\edef\csname barcolor.#1\endcsname{\pgfplotsretval}
					\pgfplotstablegetelem{#1}{Start}\of{\datatable}
					\expandafter\edef\csname barstart.#1\endcsname{\pgfplotsretval}
					\pgfplotstablegetelem{#1}{Duration}\of{\datatable}
					\expandafter\edef\csname barduration.#1\endcsname{\pgfplotsretval}
					\draw [draw=\csname barcolor.#1\endcsname,line width=6pt] (axis cs:{\csname barstart.#1\endcsname,#1}) -- (axis cs:{\csname barstart.#1\endcsname+\csname barduration.#1\endcsname,#1});
				}
			\end{axis}
		\end{tikzpicture}
		\caption{Bare-metal+Western Digital}
	\end{subfigure}%
	\hfill%
	\begin{subfigure}{0.24\textwidth}%
		\centering
		\begin{tikzpicture}
			\pgfplotstableread{
				i  Module     Start     Duration Color
				0  ROS        0         0.721432 Sepia
				1  Vehicle    0.920541  2.979571 Peach
				2  System     0.966967  3.599884 SpringGreen
				3  Map        1.640194  3.512133 Thistle
				4  Sensing    1.713815  4.517538 MidnightBlue
				5  Location   2.070260  3.882537 Mahogany
				6  Perception 0.721432  6.669958 Cerulean
				7  Planning   2.971566  3.255337 Dandelion
				8  Control    3.363737  1.496443 Periwinkle
				9  Api        3.434246  2.619346 Rhodamine
				10 Rviz       3.793541  2.123946 Emerald
				11 Total      0         7.391390 lightgray
			}\datatable
			\pgfplotstablegetrowsof{\datatable}
			\pgfmathsetmacro{\tablerows}{int(\pgfplotsretval-1)}
			\begin{axis}[
				xbar stacked,
				axis x line=bottom,
				axis y line=none,
				bar width=6pt,
				every axis plot/.append style={fill,draw=none,no markers},
				nodes near coords,
				nodes near coords align={right},
				nodes near coords style={/pgf/number format/.cd,fixed zerofill,precision=2},
				point meta=explicit,
				xmajorgrids,
				xtick style={draw=none},
				ymax=11,
				ymin=0,
				y dir=reverse,
				yticklabels from table={\datatable}{Module},
				ytick style={draw=none},
				minimum/.style={forget plot,draw=none,fill=none},
				select row/.style={
					x filter/.code={\ifnum\coordindex=#1\else\def\pgfmathresult{}\fi}
				},
				]
				\addplot [minimum,nodes near coords align={left},point meta=explicit symbolic] table [x=Start,y=i,meta=Module] {\datatable};
				\addplot +[draw=none,fill=none] table [x=Duration,y=i,meta=Duration] {\datatable};
				\pgfplotsinvokeforeach{0,...,\tablerows}{
					\pgfplotstablegetelem{#1}{Color}\of{\datatable}
					\expandafter\edef\csname barcolor.#1\endcsname{\pgfplotsretval}
					\pgfplotstablegetelem{#1}{Start}\of{\datatable}
					\expandafter\edef\csname barstart.#1\endcsname{\pgfplotsretval}
					\pgfplotstablegetelem{#1}{Duration}\of{\datatable}
					\expandafter\edef\csname barduration.#1\endcsname{\pgfplotsretval}
					\draw [draw=\csname barcolor.#1\endcsname,line width=6pt] (axis cs:{\csname barstart.#1\endcsname,#1}) -- (axis cs:{\csname barstart.#1\endcsname+\csname barduration.#1\endcsname,#1});
				}
			\end{axis}
		\end{tikzpicture}
		\caption{Bare-metal+Intel}
	\end{subfigure}%
	\caption{Comparison of Start-up time (in seconds) under disk w/ or w/o fsync optimization}
	\label{fig:start-up-disk}
\end{figure}

The selection of the disk can be influenced by applications that frequently use fsync; however, it is not evident whether Autoware consistently employs fsync.
Hence, to demonstrate its influence on launch performance, we have chosen two distinct disks for evaluation: the Intel disk and the Western Digital disk.
As shown in Fig.~\refblue{fig:start-up-disk}, the Start-up performance of the Intel and Western Digital disks is identical in this context.
This suggests that fsync optimization does not enhance the start-up times within the Autoware framework.
Therefore, the Western Digital disk emerges as the preferable option, achieving optimal speeds without the need for fsync optimization.
\subsubsection{Choice of Container}
Docker is selected over other container engines, such as Systemd-Nspawn, which lacks GPU support, and Podman, which has less mature GPU support. 
We select k3s~(v1.25.3), a lightweight version of Kubernetes, to orchestrate multiple containers.
The comparison of k3s and other lightweight version of Kubernetes is shown in Table~\refblue{table: k8sdiff}.
Additionally, we have configured k3s to use Docker as the default container runtime instead of containerd.
We employ the nvidia-docker2 package to facilitate GPU management within Docker and the nvidia-device-plugin for GPU integration with k3s.
KVM is configured in the following way: each VM is assigned with \SI{28000}{\mega\byte} RAM, and \SI{150}{\giga\byte} disk space and number of vCPUs is set to match the platform, with 24 vCPUs for the workstation and 32 vCPUs for AVA.
The \emph{native} I/O mechanism and cache option \emph{none} are set for the disk to optimize performance.
The VMs use a GPU pass-through mechanism for direct access to the Nvidia GPU, while the host system operates on the integrated GPU.

\begin{table}[tb]
	\caption{Comparison of Kubernetes, k3s, and Microk8s}
	\centering
	\label{table: k8sdiff}
	\begin{tabular}{lp{2.1cm}p{2cm}p{1.5cm}}
		\toprule
		Feature & Kubernetes & k3s & Microk8s \\
		\midrule
		Setup & Complex & Simplified & Simplified \\
		Min memory & \SI{2}{\giga\byte} & \SI{512}{\mega\byte} & \SI{1}{\giga\byte} \\
		Runtime & Docker, Containerd, CRI-O, others & Docker, Containerd, CRI-O & containerd, kata\\
		Ecosystem & Extensive &  No add-ons & k8s focused \\
		\bottomrule
	\end{tabular}
\end{table}

\subsection{Generic Performance Experiments}
The purpose of the first experiment is to evaluate the effect on generic performance of deploying the Autoware framework in containers or VMs as well as integrating enhanced microservice architecture.
\subsubsection{Evaluation metrics}
In this section, we assess the CPU and memory usage, as well as the execution time, by replaying the rosbag recorded with sensory data from an actual vehicle.  
The Autoware framework processes this data to navigate the vehicle within a simulated environment. 
In this way, we can repeat the experiment multiple times to get a stable and reliable result.
Throughout the experiment, we continuously monitor the CPU and memory utilization of each module. 
The final reported metrics represent the average values calculated over the duration of the rosbag replay.
We tested both docker and KVM in the experiments.

The experiment setups include bare-metal, docker with a single container, KVM with a single VM, and k3s with multiple containers~(10 for modular level, 17 and 26 for function level). 
The results are the average values across ten repeated runs. 
\subsubsection{Experiment results}
The CPU and RAM utilization details are listed in Table~\refblue{table: util}, where the terms \emph{k3s(10)}, \emph{k3s(17)}, and \emph{k3s(26)} represent different levels of microservice architecture: module level with ten containers, and function level with 17 and 26 containers, respectively.
The abbreviations \emph{KD} and \emph{KK} refer to Docker under KVM and k3s with 10 containers under KVM. 
The highest and lowest values in the table are highlighted in \textcolor{mycolor6}{red} and \textcolor{mycolor4}{green}.
From the result of amd64 platform, we observe that the mean CPU usage in the KVM setup is 23\% higher than that in bare-metal configurations, whereas other setups show 3 to 10\% lower CPU usage.
The term \emph{Std} represents the standard deviation, indicating the stability of the data.
The function-level architecture with 26 containers~(\emph{k3s(26)}) exhibits the lowest standard deviation, denoting high stability, while \emph{KD} and \emph{KK} show higher deviations than the bare-metal setup.
In terms of maximum CPU usage, \emph{k3s(26)} records the lowest, whereas \emph{KK} has the highest.

Regarding RAM utilization, all configurations have a higher mean utilization compared to bare-metal. 
Docker shows the lowest mean RAM usage, while \emph{KK} has the highest.
In terms of stability, indicated by the \emph{Std} value, \emph{k3s(17)} demonstrates the most stable performance, while \emph{KD} and \emph{KK} are more volatile compared to bare-metal. 
Consistent with the mean values, Docker and \emph{KK} show the lowest and highest maximal RAM usage, respectively.

The results from the ARM64 platform exhibit similar trends. 
The k3s setups consistently show lower values in all CPU-related metrics compared to the KVM setup, and the standard deviation of the k3s setups is also smaller than that of both the bare-metal and KVM setups.
Regarding RAM utilization, all the virtualized setups demonstrate higher mean and maximum values. 
Contrary to the results observed on the AMD64 platform, Docker on the ARM64 platform shows significantly higher RAM usage, approximately 20\% more than the bare-metal setup. 
Additionally, the RAM usage of KVM is also higher, by about 10\%.

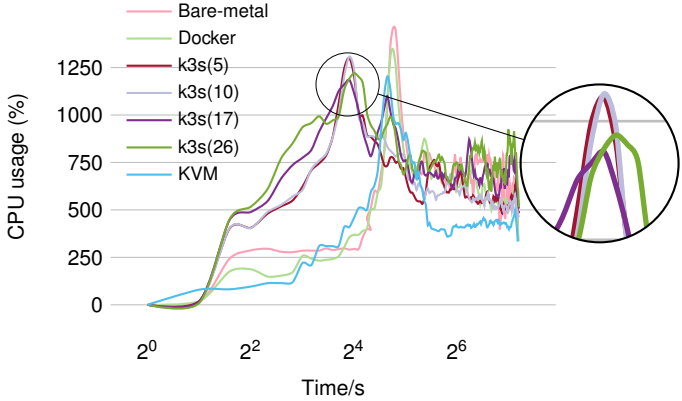
\begin{figure}[t]
	\centering
	\sffamily\sansmath
	\footnotesize
	\begin{tikzpicture}[spy using outlines={circle, magnification=2.5, size=2.1cm, connect spies}]
		\begin{axis}[
			xlabel=Time/s,
			ylabel=CPU usage~(\%),
			ymajorgrids,
			y tick label style={/pgf/number format/1000 sep=},
			x tick label style={/pgf/number format/1000 sep=},
			legend cell align={left},
			height=6cm,
			width=7.5cm,
			xmode=log,
			log basis x=2,
			axis line style={draw=none},
			legend style={font=\scriptsize, legend columns=1, at={(0.2,1.025)},anchor=north, draw=none, fill=none},
			ytick={0,250,500,750,1000,1250},
			xtick style={draw=none},
			ytick style={draw=none},
			]
			\addplot[smooth, thick, color=mycolor0] table[x=Time, y=CPU] {data/arm-bare-system-usage-processed-with-zeros.dat};
			\addlegendentry{Bare-metal}
			\addplot[smooth, thick, color=mycolor1] table[x=Time, y=CPU] {data/arm-docker-system-usage-processed-with-zeros.dat};
			\addlegendentry{Docker}
			\addplot[smooth, thick, color=mycolor6] table[x=Time, y=CPU] {data/arm-5-k3s-system-usage-processed-with-zeros.dat};
			\addlegendentry{k3s(5)}
			\addplot[smooth, thick, color=mycolor2] table[x=Time, y=CPU] {data/arm-10-k3s-system-usage-processed-with-zeros.dat};
			\addlegendentry{k3s(10)}
			\addplot[smooth, thick, color=mycolor3] table[x=Time, y=CPU] {data/arm-17-k3s-system-usage-processed-with-zeros.dat};
			\addlegendentry{k3s(17)}
			\addplot[smooth, thick, color=mycolor4] table[x=Time, y=CPU] {data/arm-26-k3s-system-usage-processed-with-zeros.dat};
			\addlegendentry{k3s(26)}
			\addplot[smooth, thick, color=mycolor5] table[x=Time, y=CPU] {data/first-arm-kvm-system-usage.dat};
			\addlegendentry{KVM}
			\spy on (3.15,3.3) in node at (6.5,2.25);
		\end{axis}	
	\end{tikzpicture}
	\caption{Autoware CPU utilization by time under different setups (arm64), 100\% means one CPU core is fully used.}
	\label{fig:CPU-time-arm}
\end{figure}

\begin{figure}[t]
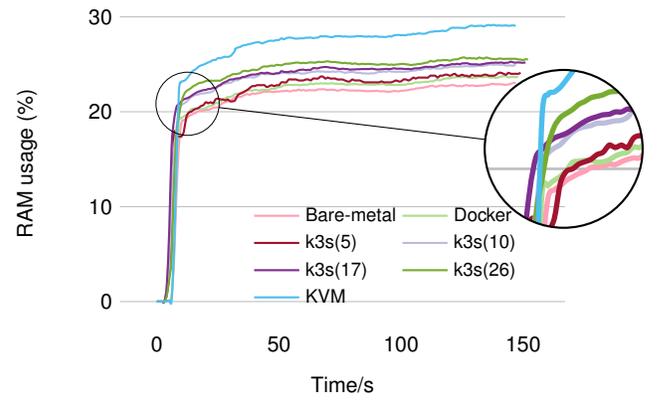

	\centering
	\sffamily\sansmath
	\footnotesize

	\caption{Autoware RAM utilization by time under different setups.}
	\label{fig:RAM-time}
\end{figure}

Moreover, we provide a more intuitive results by illustrating the changes in CPU and RAM usage over time in Figs.~\refblue{fig:CPU-time},~\refblue{fig:CPU-time-arm} ~\refblue{fig:RAM-time}, and~\refblue{fig:RAM-time-arm}, respectively.
Fig.~\refblue{fig:CPU-time} and~\refblue{fig:CPU-time-arm} reveals that the peak CPU usage for \emph{k3s(10)} and \emph{k3s(5)} occurs much earlier compared to the bare-metal setup while \emph{k3s(26)}'s occurs relatively later.
This is exactly the same with the performance of start-up time. 
The trends for Docker and KVM closely mirror those of the bare-metal, while the k3s setups are similar to each other. 
As for RAM usage, all setups display comparable trends, with the only distinction being the total amount of RAM utilized. 


\begin{table*}[tb]
	\caption{CPU and RAM utilization of single container and microservice setups. Numbers specify percentages compared to the bare-metal reference~(\SI{100}{\percent}). Minimal and maximal value are marked in \textcolor{mycolor4}{green} and \textcolor{mycolor6}{red}. \emph{K+D} and \emph{K+k} stand for Docker and k3s under KVM. \emph{k3s(x)} stand for k3s managing x containers} 
	\centering
	\label{table: util}
	\sisetup{group-digits=false,table-number-alignment=right,round-mode=places,round-precision=2,table-format=3.2}%
	\begin{tabular}{p{0.4cm} S S S S S S S S S S S S S S}
		\toprule
		&\multicolumn{7}{c}{{CPU (amd64)}} & \multicolumn{7}{c}{{RAM (amd64)}} \\
		\cmidrule(lr){2-8}
		\cmidrule(lr){9-15}
		& {Docker} & {k3s(10)} & {k3s(16)} & {k3s(26)} & {KVM} & {K+D} & {K+k} & {Docker} & {k3s(10)} & {k3s(16)} & {k3s(26)} & {KVM} & {K+D} & {K+k} \\ 
		\midrule
		Mean & 93.07 & \color{mycolor4}92.73 & 93.84 & 98.28 
		& \color{mycolor6}123.42 & 92.84 & 94.67 & \color{mycolor4}102.79 & 109.77 & 111.61
		& 113.45 & 127.37 & 123.19 & \color{mycolor6}132.12
		\\
		Std & 97.18 & 97.49 & 99.98 & \color{mycolor4}97.02 
		& 97.95& 102.77 & \color{mycolor6}104.50 & 101.88 & 95.34& \color{mycolor4}89.53
		& 101.84 & 107.42 & \color{mycolor6}129.65 & 122.61
		\\
		Max & 100.59 & 115.75 & 95.67 & \color{mycolor4}87.48 
		& 101.08 & 97.15 & \color{mycolor6}107.60 &\color{mycolor4}103.73& 108.83& 110.33
		& 112.24 & 126.91 & 124.48 & \color{mycolor6}132.03 
		\\
		\midrule
		&\multicolumn{7}{c}{{CPU (arm64)}} & \multicolumn{7}{c}{{RAM (arm64)}} \\
		\cmidrule(lr){2-8}
		\cmidrule(lr){9-15}
		& {Docker} & {k3s(10)} & {k3s(16)} & {k3s(26)} & {KVM} & {K+D} & {K+k} & {Docker} & {k3s(10)} & {k3s(16)} & {k3s(26)} & {KVM} & {K+D} & {K+k} \\ 
		\midrule
		Mean & 99.23 & \color{mycolor4}98.31 & 103.56 & 108.67 
		& \color{mycolor6}113.32 & 102.84 & 103.11 & 122.79 & 105.61 & \color{mycolor4}103.93
		& 109.69 & 124.56 & 124.85 & \color{mycolor6}148.33 
		\\
		Std & 99.61 & {93.62} & \color{mycolor4}92.18 & 95.68 
		& 99.67 & \color{mycolor6}102.87 & 100.76 & 122.08 & 91.65& \color{mycolor4}97.89
		& 101.50 & \color{mycolor6}149.38 & 139.22 & 130.98
		\\
		Max & 120.61 & 105.46 & 102.33 & \color{mycolor4}97.11
		& 99.98 & 100.67 & \color{mycolor6}109.62 & 138.52 & \color{mycolor4}105.83& 109.32
		& 110.04 & 142.72 & 144.45 & \color{mycolor6}146.22
		\\
		\bottomrule
	\end{tabular}
\end{table*}

\begin{figure}[t]
	\centering
	\sffamily\sansmath
	\footnotesize
	\begin{tikzpicture}[spy using outlines={circle, magnification=2.5, size=2.1cm, connect spies}]
		\begin{axis}[
			xlabel=Time/s,
			ylabel=CPU usage~(\%),
			ymajorgrids,
			y tick label style={/pgf/number format/1000 sep=},
			x tick label style={/pgf/number format/1000 sep=},
			legend cell align={left},
			height=6cm,
			width=7.5cm,
			xmode=log,
			log basis x=2,
			axis line style={draw=none},
			legend style={font=\scriptsize, legend columns=1, at={(0.2,1)},anchor=north, draw=none, fill=none},
			ytick={0,500,1000,1500,2000,2500},
			xtick style={draw=none},
			ytick style={draw=none},
			]
			\addplot[smooth, thick, color=mycolor0] table[x=second, y=CPU] {data/wcm-bare-metal-merged.dat};
			\addlegendentry{Bare-metal}
			\addplot[smooth, thick, color=mycolor1] table[x=second, y=CPU] {data/wcm-docker-rosbag-merged.dat};
			\addlegendentry{Docker}
			\addplot[smooth, thick, color=mycolor6] table[x=second, y=CPU] {data/k3s-rosbag-5-merged.dat};
			\addlegendentry{k3s(5)}
			\addplot[smooth, thick, color=mycolor2] table[x=second, y=CPU] {data/wcm-k3s-rosbag-10-merged.dat};
			\addlegendentry{k3s(10)}
			\addplot[smooth, thick, color=mycolor3] table[x=second, y=CPU] {data/wcm-k3s-rosbag-17-merged.dat};
			\addlegendentry{k3s(17)}
			\addplot[smooth, thick, color=mycolor4] table[x=second, y=CPU] {data/wcm-k3s-rosbag-26-merged.dat};
			\addlegendentry{k3s(26)}
			\addplot[smooth, thick, color=mycolor5] table[x=second, y=CPU] {data/kvm-bare-metal-merged.dat};
			\addlegendentry{KVM}
			\spy on (2.4,3.7) in node at (5.5,2.25);
		\end{axis}
	\end{tikzpicture}
	\caption{Autoware CPU utilization by time under different setups, 100\% means one CPU core is fully used.}
	\label{fig:CPU-time}
\end{figure}

\subsection{Start-Up Time in Container/VM}
This section is dedicated to evaluating the impact of containerization, virtualization, and microservice architecture on the start-up time of the Autoware framework.
\subsubsection{Evaluation metric}
Firstly, it is essential to establish a clear definition for the start-up time of the Autoware framework.
Autoware operates with 242 ROS nodes, which encompass both standard and composable nodes.
These nodes are initiated through Python scripts and C++ code. 
The utilization of composable nodes is strategically aimed at facilitating the execution of multiple nodes within a single process to minimize overhead and potentially enhance communication efficiency.
Modifying the source code of Autoware and ROS makes it possible to record the initiation and load completion times of the various nodes.
The execution of the \emph{ros2 launch} command marks the commencement time, while the completion of the last node's loading signifies the end time.
This methodology enables the determination of the start-up time for the entire application as well as for individual modules.

\begin{figure}[t]
	\centering
	\sffamily\sansmath
	\footnotesize
	%
	\caption{Autoware RAM utilization by time under different setups (arm64).}
	\label{fig:RAM-time-arm}
\end{figure}

\subsubsection{Start-up Time in Containerization}

\begin{figure*}[]
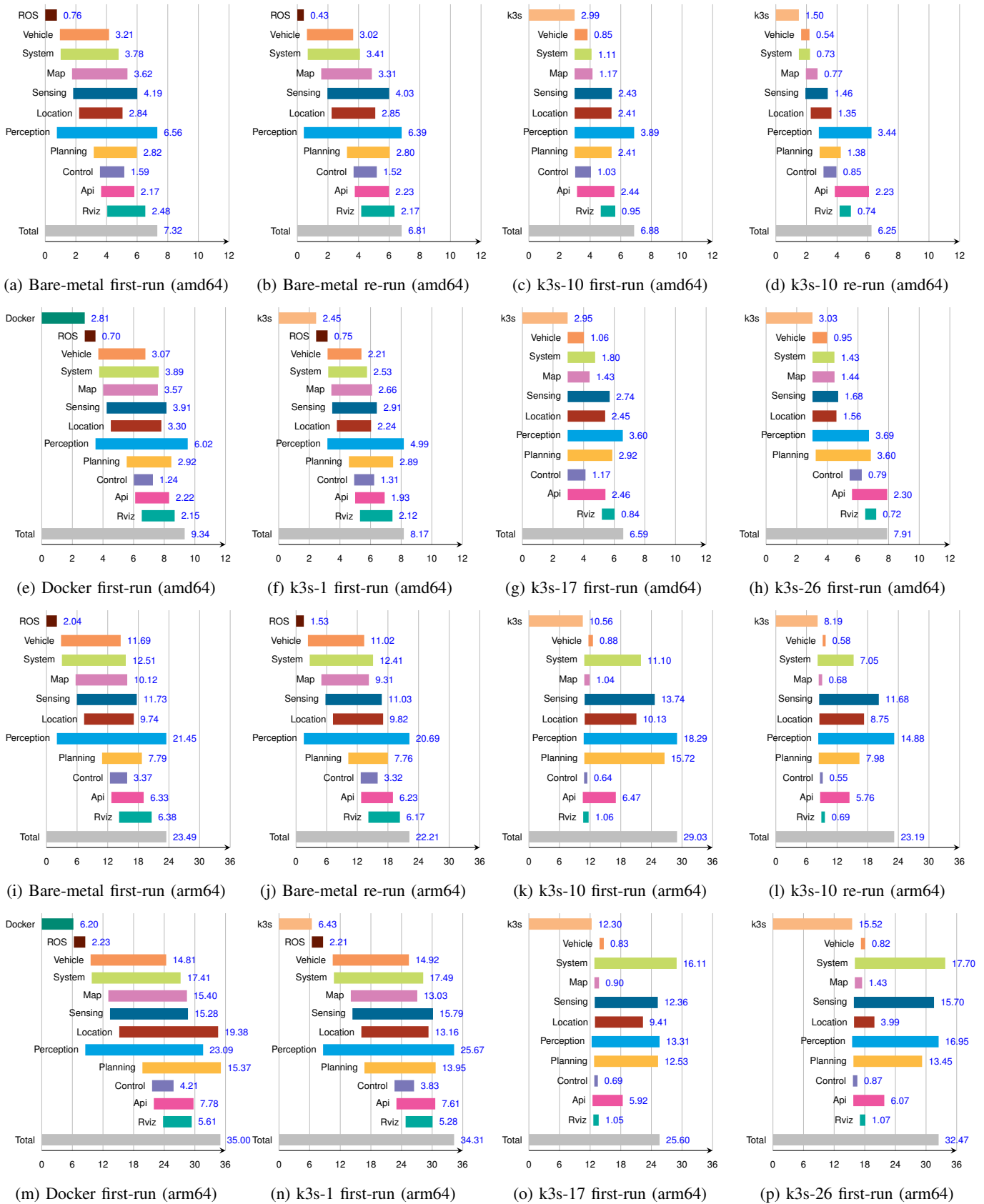

	\pgfplotsset{
		enlarge y limits=0.05,
		every axis/.append style={font=\sansmath\sffamily\tiny},
		xlabel={},
		every axis x label/.style={at={(ticklabel* cs:1.0)},anchor=south east},
		height=6cm,
		width=5cm,
		xmax=12,
		xtick={0,2,4,6,8,10,12},
		ytick={0,1,2,3,4,5,6,7,8,9,10,11,12},
	}
	\begin{subfigure}{0.24\textwidth}%
		\centering
		%
		\caption{Bare-metal first-run (amd64)}
		\label{figure:6a}
	\end{subfigure}%
	\hfill%
	\begin{subfigure}{0.24\textwidth}
		\centering
		%
		\caption{Bare-metal re-run (amd64)}
		\label{figure:6b}
	\end{subfigure}%
	\hfill%
	\begin{subfigure}{0.24\textwidth}
		\centering
		%
		\caption{k3s-10 first-run (amd64)}
		\label{figure:6c}
	\end{subfigure}%
	\hfill%
	\begin{subfigure}{0.24\textwidth}%
		\centering
		%
		\caption{k3s-10 re-run (amd64)}
		\label{figure:6d}
	\end{subfigure}%
	\medskip%
	\\%
	\begin{subfigure}{0.24\textwidth}%
		\centering
		%
		\caption{Docker first-run (amd64)}
		\label{figure:6e}
	\end{subfigure}%
	\hfill%
	\begin{subfigure}{0.24\textwidth}%
		\centering
		%
		\caption{k3s-1 first-run (amd64)}
		\label{figure:6f}
	\end{subfigure}%
	\hfill%
	\begin{subfigure}{0.24\textwidth}%
		\centering
		%
		\caption{k3s-17 first-run (amd64)}
		\label{figure:6g}
	\end{subfigure}%
	\hfill%
	\begin{subfigure}{0.24\textwidth}%
		\centering
		%
		\caption{k3s-26 first-run (amd64)}
		\label{figure:6h}
	\end{subfigure}%
	\hfill%
	\medskip%
	\\%
	\begin{subfigure}{0.24\textwidth}%
		\centering
		%
		\caption{Bare-metal first-run (arm64)}
		\label{figure:6i}
	\end{subfigure}%
	\hfill%
	\begin{subfigure}{0.24\textwidth}
		\centering
		%
		\caption{Bare-metal re-run (arm64)}
		\label{figure:6j}
	\end{subfigure}%
	\hfill%
	\begin{subfigure}{0.24\textwidth}
		\centering
		%
		\caption{k3s-10 first-run (arm64)}
		\label{figure:6k}
	\end{subfigure}%
	\hfill%
	\begin{subfigure}{0.24\textwidth}%
		\centering
		%
		\caption{k3s-10 re-run (arm64)}
		\label{figure:6l}
	\end{subfigure}%
	\medskip%
	\\%
	\begin{subfigure}{0.24\textwidth}%
		\centering
		%
		\caption{Docker first-run (arm64)}
		\label{figure:6m}
	\end{subfigure}%
	\hfill%
	\begin{subfigure}{0.24\textwidth}%
		\centering
		%
		\caption{k3s-1 first-run (arm64)}
		\label{figure:6n}
	\end{subfigure}%
	\hfill%
	\begin{subfigure}{0.24\textwidth}%
		\centering
		%
		\caption{k3s-17 first-run (arm64)}
		\label{figure:6o}
	\end{subfigure}%
	\hfill%
	\begin{subfigure}{0.24\textwidth}%
		\centering
		%
		\caption{k3s-26 first-run (arm64)}
		\label{figure:6p}
	\end{subfigure}%
	\caption{Start-up time (in seconds) of modules across different configurations and platforms (amd64 and arm64). \emph{k3s-1(10, 17, 26)} stands for dividing the Autoware framework into 1 (10, 17, 26) segment(s) and using k3s to launch them.}
	\label{fig:start-up} 
\end{figure*}

Fig.~\refblue{fig:start-up} delineates the start-up time for the entire Autoware framework and its various modules under eight distinct scenarios.
Fig.~\refblue{figure:6c} presents the results when Autoware is integrated with a function-level microservice architecture, separated into ten containers orchestrated by k3s.
The data indicates that k3s requires approximately \SI{3.0}{\second} to initiate all containers.
With the exception of Rviz, all modules are launched simultaneously.
Within this figure, the bar labeled \emph{k3s} represents the aggregate time consumed by k3s for deploying as well as for the container creation process, which will be referred to as \emph{k3s launch time}.
Rviz is a visualization tool of ROS. 
Rviz, functioning as ROS's visualization tool, demands preliminary processing to relay visualized data within its container, thereby resulting in a delayed launch relative to other modules. 
Notably, the perception module experiences the longest start-up time, which is anticipated considering its reliance on the GPU to load an extensive array of libraries essential for CUDA acceleration. 
In this scenario, the application's start-up is completed in 6.9 seconds.

Fig.~\refblue{figure:6a} illustrates the start-up times under the bare-metal configuration.
Unlike the results of k3s, modules are launched at different times on bare-metal.
Diverging from the k3s, the bare-metal setup launches modules one after the other instead of simultaneously.
Surprisingly, the start-up time for bare-metal exceeds that of k3s by \SI{0.4}{\second}, even when considering k3s's own overhead~(deploying and container creation). 
To delve deeper, we conducted further experiments by deploying the entire application within a single container and utilizing both Docker and k3s, as detailed in Figs.~\refblue{figure:6e} and~\refblue{figure:6f}. 
The experiments reveal launch times of \SI{2.8}{\second} for Docker and \SI{2.4}{\second} for k3s. 
The initialization time of ROS remains consistent across bare-metal, Docker, and k3s, at around \SI{0.7}{\second}.
The overall start-up time is longer than that observed in the bare-metal configuration, with Docker being notably slower.
Yet, if the time consumed by Docker and k3s for container creation is excluded, Autoware's start-up proves to be marginally quicker than on bare-metal.
Consequently, it is the division into separate modules that provide significant performance improvement.
The benefit of dividing exceeds the container's overhead, thus resulting in a better performance compared to bare-metal.
 
Upon examining Figs.~\refblue{figure:6a} and~\refblue{figure:6c}, it becomes evident that module launch times are significantly reduced with k3s;
in fact, when applying the enhanced microservice architecture, the modules launch in nearly half the time compared to the bare-metal setup.
This observation does not mean the enhanced microservice architecture performs much better than bare-metal.
The enhanced efficiency primarily stems from the modular containerization in k3s, where each module operates within its dedicated container.
We elaborate further on this with the vehicle module as an example.
As shown in Table~\refblue{table: module}, this module comprises a single node: the robot-state-publisher.
This ROS node parses various parameters to establish a connection between the vehicle and its sensors.  
Within the Autoware framework, certain dependent nodes must be initialized prior to the main nodes.
For the vehicle module operating in the vehicle container in k3s, the robot-state-publisher is the sole dependent node.
Once it is launched, the module commences the synchronization of vehicle information with the corresponding sensors.
Once this procedure is done, the vehicle module's initialization is considered fully accomplished.

Conversely, in a bare-metal setup, ROS is tasked with launching a suite of ten modules at one time.
It has to launch all associated dependent nodes first, which far exceeds the number for a single module.
For the vehicle module on bare-metal, it has to launch all 85 dependent nodes before it can proceed with the task after launching the robot-state-publisher node.
Therefore, the vehicle module's launch time in a bare-metal environment includes the time taken to launch these dependent nodes.
This is the reason why modules have extended launch times on bare-metal. 

As previously emphasized, it is critical for automotive applications to possess short (re-)start times to minimize both boot time and recovery intervals after a software crash. 
It is worthwhile to assess the influence of first-run versus re-run. The \emph{first-run} refers to the operation of the application for the first time after boot, whereas \emph{re-run} denotes the operation of the application after it has already been executed multiple times.
The previously discussed results were all obtained after a system reboot; hence, for a comprehensive analysis, the outcomes of the re-run have been included for comparison in Figs.~\refblue{figure:6b}, and \refblue{figure:6d}.
Fig.~\refblue{figure:6d} shows a reduction in the launch time for k3s, decreasing from \SI{3.0}{\second} to \SI{1.5}{\second}, with a corresponding decrease observed across all modules.
However, the modules no longer initiate simultaneously, which accounts for the smaller decline in total launch time compared to the \SI{1.5}{\second} saved by k3s.
The second notable distinction is the launch time of ROS. 
The data in Figs.~\refblue{figure:6b} indicate that the ROS launch time reduced by approximately \SI{0.3}{\second} compared to first-run.
This reduction contributes to the decreased launch time of all modules observed in Fig.~\refblue{figure:6d}, as each container must independently initiate ROS.
Figs.~\refblue{figure:6e} and~\refblue{figure:6f} illustrate that using k3s to launch a single container containing the entire Autoware framework results in faster launch times compared to using Docker to launch Autoware. 

To deepen our understanding of the impact of microservice architecture, we conducted a detailed investigation focusing on the function-level architecture.
Our first test involved a function-level architecture comprising 17 containers, within which we specifically separate the perception and planning modules.
Our findings, as illustrated in Fig.~\refblue{figure:6g}, reveal that the first-run start-up time of this architecture is comparable to that of the module-level architecture.
Specifically, the start-up time of the function-level architecture is approximately 4.5\% (or \SI{0.28}{\second}) faster than the module-level counterpart.
If we dive deeper into the launch time of perception and planning module which we further separate in this experiment.
We can notice that the perception module launches faster compared to the module-level architecture, whereas the planning module takes longer.
This investigation suggests that the further divide of modules in a microservice architecture does not consistently lead to reduced launch times.

The outcomes of the function-level structure, which incorporates 26 containers, are presented in Figs.~\refblue{figure:6h}.
It is evident from these results that the start-up time for this configuration is longer compared to the bare-metal setup.
Additionally, it was observed that modules, specifically control, API, and RViz, experienced a delayed start in this setup.

Figs.~\refblue{figure:6i} -~\refblue{figure:6p} present the start-up times on the ARM64 platform with the same experimental configuration.
Autoware launches much slower on the ARM64 platform due to hardware differences, but we can draw some similar conclusions from the results.
First, Docker still launches Autoware the slowest among all the setups.
Second, the function-level architecture with 17 containers launches faster than the one with 10 containers, with a greater improvement in launch time (13.1\%) compared to the AMD64 platform (4.5\%).
Similarly, the decrease in launch time for the function-level architecture with 26 containers is smaller (11.8\%) than on the AMD64 platform (20.0\%). 
This outcome aligns with the fact that the ARM64 platform has more CPU cores.

A notable difference between these two platforms is that on the ARM64 platform, all the k3s setups take longer to launch Autoware compared to the bare-metal setup. 
This is primarily due to the significantly longer start-up time of k3s; it takes more than twice the time to launch multiple containers compared to launching a single container, which only takes slightly longer (about 20\%) on the AMD64 platform.
However, the start-up time for the function-level architecture with 17 containers is very close to that of the bare-metal setup, indicating the feasibility of this enhanced microservice architecture on the ARM64 platform.

\begin{figure*}[t]
	\centering
	\pgfplotsset{
		enlarge y limits=0.05,
		every axis/.append style={font=\sansmath\sffamily\tiny},
		xlabel={},
		every axis x label/.style={at={(ticklabel* cs:1.0)},anchor=south east},
		height=6cm,
		width=5cm,
		xmax=12,
		xtick={0,2,4,6,8,10,12},
		ymax=12,
		ymin=0,
		ytick={0,1,2,3,4,5,6,7,8,9,10,11,12},
	}
	\begin{subfigure}{0.24\textwidth}%
		\centering
		\begin{tikzpicture}
			\pgfplotstableread{
				i  Module     Start     Duration Color
				0  k3s        0.000     2.929    Apricot
				1  Vehicle    3.668     0.838    Peach
				2  System     3.566     1.017    SpringGreen
				3  Map        2.929     1.229    Thistle
				4  Sensing    2.929     3.208    MidnightBlue
				5  Location   2.942     3.122    Mahogany
				6  Perception 3.045     3.869    Cerulean
				7  Planning   2.929     1.882    Dandelion
				8  Control    3.569     1.147    Periwinkle
				9  Api        2.929     3.36     Rhodamine
				10 Rviz       4.825     0.936    Emerald
				11 Total      0         6.913    lightgray 
			}\datatable
			\pgfplotstablegetrowsof{\datatable}
			\pgfmathsetmacro{\tablerows}{int(\pgfplotsretval-1)}
			\begin{axis}[
				xbar stacked,
				axis x line=bottom,
				axis y line=none,
				bar width=6pt,
				every axis plot/.append style={fill,draw=none,no markers},
				nodes near coords,
				nodes near coords align={right},
				nodes near coords style={/pgf/number format/.cd,fixed zerofill,precision=2},
				point meta=explicit,
				xmajorgrids,
				xtick style={draw=none},
				ymax=11,
				ymin=0,
				y dir=reverse,
				yticklabels from table={\datatable}{Module},
				ytick style={draw=none},
				minimum/.style={forget plot,draw=none,fill=none},
				select row/.style={
					x filter/.code={\ifnum\coordindex=#1\else\def\pgfmathresult{}\fi}
				},
				]
				\addplot [minimum,nodes near coords align={left},point meta=explicit symbolic] table [x=Start,y=i,meta=Module] {\datatable};
				\addplot +[draw=none,fill=none] table [x=Duration,y=i,meta=Duration] {\datatable};
				\pgfplotsinvokeforeach{0,...,\tablerows}{
					\pgfplotstablegetelem{#1}{Color}\of{\datatable}
					\expandafter\edef\csname barcolor.#1\endcsname{\pgfplotsretval}
					\pgfplotstablegetelem{#1}{Start}\of{\datatable}
					\expandafter\edef\csname barstart.#1\endcsname{\pgfplotsretval}
					\pgfplotstablegetelem{#1}{Duration}\of{\datatable}
					\expandafter\edef\csname barduration.#1\endcsname{\pgfplotsretval}
					\draw [draw=\csname barcolor.#1\endcsname,line width=6pt] (axis cs:{\csname barstart.#1\endcsname,#1}) -- (axis cs:{\csname barstart.#1\endcsname+\csname barduration.#1\endcsname,#1});
				}
			\end{axis}
		\end{tikzpicture}
		\caption{k3s-5 first-run (amd64)}
		\label{figure:7a}
	\end{subfigure}%
	\hfill%
	\begin{subfigure}{0.24\textwidth}%
		\centering
		\begin{tikzpicture}
			\pgfplotstableread{
				i  Module     Start     Duration Color
				0  k3s        0.000     10.760   Apricot
				1  Vehicle    11.588    0.8322   Peach
				2  System     11.002    11.123   SpringGreen
				3  Map        11.157    1.178    Thistle
				4  Sensing    11.272    13.554   MidnightBlue
				5  Location   11.303    10.201   Mahogany
				6  Perception 10.946    18.655   Cerulean
				7  Planning   11.040    15.583   Dandelion
				8  Control    11.077    0.707    Periwinkle
				9  Api        10.760     2.027    Rhodamine
				10 Rviz       10.933     0.772    Emerald
				11 Total      0         29.601    lightgray 
			}\datatable
			\pgfplotstablegetrowsof{\datatable}
			\pgfmathsetmacro{\tablerows}{int(\pgfplotsretval-1)}
			\begin{axis}[
				xbar stacked,
				axis x line=bottom,
				axis y line=none,
				bar width=6pt,
				every axis plot/.append style={fill,draw=none,no markers},
				nodes near coords,
				nodes near coords align={right},
				nodes near coords style={/pgf/number format/.cd,fixed zerofill,precision=2},
				point meta=explicit,
				xmajorgrids,
				xtick style={draw=none},
				ymax=11,
				ymin=0,
				xmax=36,
				xtick={0,6,12,18,24,30,36},
				y dir=reverse,
				yticklabels from table={\datatable}{Module},
				ytick style={draw=none},
				minimum/.style={forget plot,draw=none,fill=none},
				select row/.style={
					x filter/.code={\ifnum\coordindex=#1\else\def\pgfmathresult{}\fi}
				},
				]
				\addplot [minimum,nodes near coords align={left},point meta=explicit symbolic] table [x=Start,y=i,meta=Module] {\datatable};
				\addplot +[draw=none,fill=none] table [x=Duration,y=i,meta=Duration] {\datatable};
				\pgfplotsinvokeforeach{0,...,\tablerows}{
					\pgfplotstablegetelem{#1}{Color}\of{\datatable}
					\expandafter\edef\csname barcolor.#1\endcsname{\pgfplotsretval}
					\pgfplotstablegetelem{#1}{Start}\of{\datatable}
					\expandafter\edef\csname barstart.#1\endcsname{\pgfplotsretval}
					\pgfplotstablegetelem{#1}{Duration}\of{\datatable}
					\expandafter\edef\csname barduration.#1\endcsname{\pgfplotsretval}
					\draw [draw=\csname barcolor.#1\endcsname,line width=6pt] (axis cs:{\csname barstart.#1\endcsname,#1}) -- (axis cs:{\csname barstart.#1\endcsname+\csname barduration.#1\endcsname,#1});
				}
			\end{axis}
		\end{tikzpicture}
		\caption{k3s-5 first-run (arm64)}
		\label{figure:7b}
	\end{subfigure}%
	\hfill%
	\begin{subfigure}{0.24\textwidth}%
		\centering
		\begin{tikzpicture}
			\pgfplotstableread{
				i  Module     Start     Duration	Color     
				0  ROS        0         0.849130	Sepia
				1  Vehicle    1.063550  3.730097	Peach
				2  System     1.125579  4.764525	SpringGreen
				3  Map        1.944723  4.111152	Thistle
				4  Sensing    2.020122  4.566971	MidnightBlue
				5  Location   2.453548  4.138581	Mahogany
				6  Perception 0.849130  7.258323	Cerulean
				7  Planning   3.553691  3.756742	Dandelion
				8  Control    4.022410  1.470154	Periwinkle
				9  Api        4.109526  2.538487	Rhodamine
				10 Rviz       4.574094  2.522821	Emerald
				11 Total      0         8.107453  	lightgray  
			}\datatable
			\pgfplotstablegetrowsof{\datatable}
			\pgfmathsetmacro{\tablerows}{int(\pgfplotsretval-1)}
			\begin{axis}[
				xbar stacked,
				axis x line=bottom,
				axis y line=none,
				bar width=6pt,
				every axis plot/.append style={fill,draw=none,no markers},
				nodes near coords,
				nodes near coords align={right},
				nodes near coords style={/pgf/number format/.cd,fixed zerofill,precision=2},
				point meta=explicit,
				xmajorgrids,
				xtick style={draw=none},
				ymax=11,
				ymin=0,
				y dir=reverse,
				yticklabels from table={\datatable}{Module},
				ytick style={draw=none},
				minimum/.style={forget plot,draw=none,fill=none},
				select row/.style={
					x filter/.code={\ifnum\coordindex=#1\else\def\pgfmathresult{}\fi}
				},
				]
				\addplot [minimum,nodes near coords align={left},point meta=explicit symbolic] table [x=Start,y=i,meta=Module] {\datatable};
				\addplot +[draw=none,fill=none] table [x=Duration,y=i,meta=Duration] {\datatable};
				\pgfplotsinvokeforeach{0,...,\tablerows}{
					\pgfplotstablegetelem{#1}{Color}\of{\datatable}
					\expandafter\edef\csname barcolor.#1\endcsname{\pgfplotsretval}
					\pgfplotstablegetelem{#1}{Start}\of{\datatable}
					\expandafter\edef\csname barstart.#1\endcsname{\pgfplotsretval}
					\pgfplotstablegetelem{#1}{Duration}\of{\datatable}
					\expandafter\edef\csname barduration.#1\endcsname{\pgfplotsretval}
					\draw [draw=\csname barcolor.#1\endcsname,line width=6pt] (axis cs:{\csname barstart.#1\endcsname,#1}) -- (axis cs:{\csname barstart.#1\endcsname+\csname barduration.#1\endcsname,#1});
				}
			\end{axis}
		\end{tikzpicture}
		\caption{KVM first-run (amd64)}
		\label{figure:7c}
	\end{subfigure}%
	\hfill%
	\begin{subfigure}{0.24\textwidth}%
		\centering
		\begin{tikzpicture}
			\pgfplotstableread{
				i  Module     Start     Duration	Color     
				0  ROS        0         2.347769	Sepia
				1  Vehicle    3.055144  13.362700	Peach
				2  System     3.252019  16.514887	SpringGreen
				3  Map        6.099715  14.941857	Thistle
				4  Sensing    6.363221  19.706433	MidnightBlue
				5  Location   7.817017  18.114100	Mahogany
				6  Perception 2.347769  28.046847	Cerulean
				7  Planning   11.635806 13.207654	Dandelion
				8  Control    13.280869 7.255384	Periwinkle
				9  Api        13.577146 7.704909	Rhodamine
				10 Rviz       15.164776 9.154297	Emerald
				11 Total      0         30.394616  	lightgray     
			}\datatable
			\pgfplotstablegetrowsof{\datatable}
			\pgfmathsetmacro{\tablerows}{int(\pgfplotsretval-1)}
			\begin{axis}[
				xbar stacked,
				axis x line=bottom,
				axis y line=none,
				bar width=6pt,
				every axis plot/.append style={fill,draw=none,no markers},
				nodes near coords,
				nodes near coords align={right},
				nodes near coords style={/pgf/number format/.cd,fixed zerofill,precision=2},
				point meta=explicit,
				xmajorgrids,
				xtick style={draw=none},
				ymax=11,
				ymin=0,
				xmax=36,
				xtick={0,6,12,18,24,30,36},
				y dir=reverse,
				yticklabels from table={\datatable}{Module},
				ytick style={draw=none},
				minimum/.style={forget plot,draw=none,fill=none},
				select row/.style={
					x filter/.code={\ifnum\coordindex=#1\else\def\pgfmathresult{}\fi}
				},
				]
				\addplot [minimum,nodes near coords align={left},point meta=explicit symbolic] table [x=Start,y=i,meta=Module] {\datatable};
				\addplot +[draw=none,fill=none] table [x=Duration,y=i,meta=Duration] {\datatable};
				\pgfplotsinvokeforeach{0,...,\tablerows}{
					\pgfplotstablegetelem{#1}{Color}\of{\datatable}
					\expandafter\edef\csname barcolor.#1\endcsname{\pgfplotsretval}
					\pgfplotstablegetelem{#1}{Start}\of{\datatable}
					\expandafter\edef\csname barstart.#1\endcsname{\pgfplotsretval}
					\pgfplotstablegetelem{#1}{Duration}\of{\datatable}
					\expandafter\edef\csname barduration.#1\endcsname{\pgfplotsretval}
					\draw [draw=\csname barcolor.#1\endcsname,line width=6pt] (axis cs:{\csname barstart.#1\endcsname,#1}) -- (axis cs:{\csname barstart.#1\endcsname+\csname barduration.#1\endcsname,#1});
				}
			\end{axis}
		\end{tikzpicture}
		\caption{KVM first-run (arm64)}
		\label{figure:7d}
	\end{subfigure}%
	\caption{start-up time (in seconds) of modules when Autoware is divided into 5 segments and when deployed under KVM.}
	\label{fig:start-up-5-kvm}
\end{figure*}
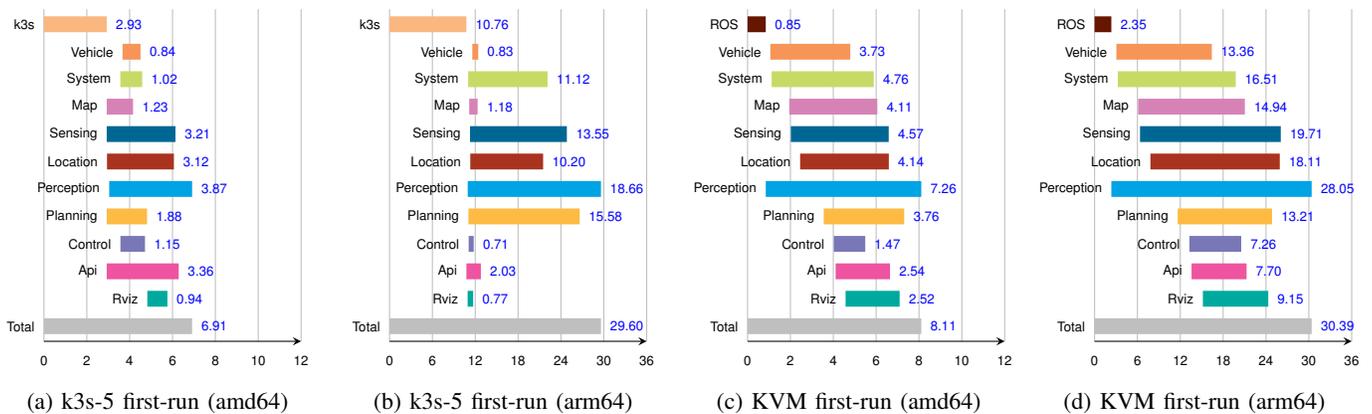

According to previous results, increasing the number of containers from 10 to 17 reduced the start-up time.
To investigate the reverse scenario, we integrate 10 modules into 5 containers: map and vehicle, api and system, sensing and perception, localization and rviz, planning and control.
The outcomes of this experiment are illustrated in Figs.~\refblue{figure:7a} and~\refblue{figure:7b}.
We observed that the start-up time in first-run is comparable to other configurations. 
However, the start-up time in re-run was marginally slower~(by approximately 4.0\%) compared to the setup with 10 containers.
Additionally, the start-up time for each individual module is longer than in the 10 container setup.
These findings indicate that although a microservice architecture can effectively reduce the application start-up time, determining the optimal number of containers requires specific testing. 
The ideal number is not necessarily the maximum or minimum possible, but rather a balance that needs to be identified through experimentation.

In conclusion, for both amd64 and arm64 platforms, running an undivided Autoware framework in a single container results in longer start-up times for both Docker and k3s compared to bare-metal, when including the launch times of Docker and k3s themselves.
Specifically, Docker's first-run start-up time is considerably slower. 
However, when the Autoware framework is restructured into a microservice architecture with not too much containers, the overall start-up time can become more efficient than on bare-metal.
Our findings suggest that containerization slightly affects Autoware's start-up time and can actually lead to improved performance with proper segmentation.

\subsubsection{Start-up Time in Virtualization}

To evaluate the impact of virtualization technology, we conducted additional experiments using KVM. 
Figs.~\refblue{figure:7c} and \refblue{figure:7d} presents the results obtained under KVM.
The observations indicate that performance on both AMD64 and ARM64 platforms declines when operating in a virtualized environment.  
The start-up time under KVM for both platforms is slower than on bare-metal and k3s, but faster than on Docker. 
Detailed results and analysis of KVM, along with re-runs of each setup, can be found in our previous work~\cite{10186789}.

\subsection{Discussion}
The previously presented results draw counter-intuitive conclusions. 
In the assessment of CPU and RAM utilization, for both amd64 and arm64 platform, containerized environments~(Docker and k3s) exhibit lower mean CPU usage along with a more stable distribution.
While these setups show a higher mean RAM usage compared to bare-metal configurations, their stability is either comparable to or greater than that of bare-metal.
This is surprising since containerization typically introduces additional overhead compared to bare-metal systems.
The experiments are repeated multiple times to ensure reliable output, and based on the existing results, we found some possible explanations for this.
Our analysis suggests that Docker containers' efficiency can be attributed to their secure, isolated environments created on the host system. 
This isolation is achieved through two primary kernel features: kernel namespaces and control groups (cgroups)~\cite{7742298}.
Kernel namespaces isolate various system aspects, limiting the ability of processes to see and interact with each other~\cite{reshetova2014security}.
Control groups, on the other hand, manage and limit the resource usage of a process or group of processes.
They prevent any single process from monopolizing available resources, thereby ensuring balanced resource allocation among all processes and containers on the host.
This balanced resource allocation, enabled by kernel namespaces and cgroups, may explain the unexpectedly efficient performance of containerized environments and microservice architecture in our experiments.
Like Docker, k3s incorporates a resource allocation and limitation mechanism within its architecture~\cite{burns2022kubernetes}. 
This feature further ensures that no single container can monopolize all available resources.

In our analysis of start-up times, we found that a single container setup is slower than a bare-metal configuration.
However, employing a function-level architecture with 17 containers results in a significant 18\%~(or \SI{0.57}{\second}) reduction in start-up time compared to the bare-metal under amd64 platform.
The number of containers required to achieve the fastest start-up time is not necessarily the highest or lowest possible. It needs to be determined through careful experimentation.
Although the start-up time using k3s on the ARM64 platform is longer than on bare-metal, it can still deliver comparable performance when factoring in the time used by k3s. 
Apart from the aforementioned factors, the number of ROS nodes could be another factor.
When the number of nodes increases, ROS's performance tends to decline, with 242 nodes constituting a substantial workload~\cite{Profanter2019}.
Segmenting the application into smaller parts with fewer nodes mitigates this performance drop.
However, this improvement in performance does not continue indefinitely with further separation.
When the total number of containers reaches 26, where each function in the module is divided into its own dedicated container, the performance declines compared to the bare-metal setup.
This phenomenon occurs on both amd64 and arm64 platform. 
At this point, hardware limitations become a significant factor; the system can no longer launch all containers simultaneously, leading to a longer start-up time.
This situation highlights a critical trade-off: maintaining optimal start-up performance versus enhancing the flexibility of the microservice architecture.
It underscores the need for a balanced approach, where the benefits of microservice architecture are weighed against potential impacts on performance.  

Figs.~\refblue{fig:CPU-time} and~\refblue{fig:CPU-time-arm} can also explain the performance improvement of microservice architecture in start-up time.
As observed in the figure, the CPU usage in microservice setups begins to rise approximately \SI{2}{\second} earlier than in bare-metal setups, this trend also extends to the point at which CPU usage reaches its peak.
This earlier increase in CPU utilization suggests a more responsive initiation of processes in microservice setups, highlighting one of its key benefits.
However, when there are too many containers that k3s can not launch them simultaneously, as shown in the figure, the resulting CPU usage trend deviates from the previously observed smooth and consistent pattern.

In summary, our benchmarks in the fields of virtualization, containerization, and microservice architecture not only reveal the flexibility and security offered by these technologies but also reveal their potential for performance enhancement for both amd64 and arm64 platforms.
Rather than merely resulting in performance loss, these technologies can actually lead to improvements when appropriately leveraged.
The factors discussed earlier are vital for achieving optimal performance in such environments.

\newcommand{\bccbar}[3][]{
	\renewcommand{\bcbarcolor}{blue!20}
	\renewcommand{\bcbartext}{}
	\renewcommand{\bcbarlabel}{}
	\renewcommand{\bcbarvalue}{#2\bcunit}
	\renewcommand{\bcplainbar}{false}
	\setkeys{bcbar}{#1}
	\fill[color=mycolor2,fill,draw] ([xshift=#3*(\bcwidth/\bcrange)]0,\bcpos) rectangle ([xshift=#3*(\bcwidth/\bcrange)]$#2-\bcmin*(\bcwidth/\bcrange,0) + (0,\bcpos-5mm)$);
	\draw ([xshift=#3*(\bcwidth/\bcrange)]0,\bcpos) rectangle ([xshift=#3*(\bcwidth/\bcrange)]$#2-\bcmin*(\bcwidth/\bcrange,0) + (0,\bcpos-5mm)$);
	\ifthenelse{\equal{\bcplainbar}{true}}{}{
		\node[anchor=west] at ([xshift=#3*(\bcwidth/\bcrange)]$#2-\bcmin*(\bcwidth/\bcrange,0) + (0,\bcpos-2.5mm)$) {\bcfontstyle\bcbarvalue};
	}
	\node[anchor=west] at ([xshift=#3*(\bcwidth/\bcrange)]0,\bcpos-2.5mm) {\bcfontstyle\bcbartext};
	\node[anchor=east] at ([xshift=#3*(\bcwidth/\bcrange)]0,\bcpos-2.5mm) {\bcfontstyle\bcbarlabel};
	\addtolength{\bcpos}{-5mm}
}

\section{Conclusion}

This paper presents performance benchmarks for hypervisors and containers, and explores the application of microservice architecture incorporated with containerization  within the context of SDVs.
We assess various virtualization and containerization technologies, such as KVM, Docker, Podman, and Nspawn, using the performance metrics of bare-metal systems as a baseline.
Our benchmarks are conducted across diverse hardware, from Raspberry Pi to high-performance AMD64 and ARM64 platforms. 
We perform a generic performance evaluation focusing on CPU, memory, network, and disk as well as the experiments under real-world scenarios employing the Autoware framework.
Furthermore, we integrate microservice architecture to the Autoware framework, separating it into different containers managed by the container orchestration tool k3s and demonstrating its feasibility.
Extensive benchmark results indicate that software operating within virtualized or containerized environments exhibit performance metrics closely aligned with those of bare-metal systems.

In generic benchmark tests, both embedded, amd64, and arm64 systems show similar outcomes. 
Software running in VMs and containers exhibits a slight performance decrease of 0-5\% in CPU, memory, and network.
Disk performance, however, is more substantially affected by virtualization and containerization, with container engines experiencing roughly a 5-15\% reduction in performance. 
Notably, disk operations within KVM deliver a 35\% loss compared to a bare-metal setup.
The results from real-world scenarios reveal that, on the AMD64 platform, start-up times for KVM and Docker are 10\% to 22\% slower than bare-metal when considering the time required for container creation.
In contrast, k3s is 18\% faster than bare-metal in cases where Autoware is integrated with an enhanced microservice architecture.
On the ARM64 platform, start-up times for k3s, KVM, and Docker are 8\%, 23\%, and 33\% slower than bare-metal, respectively.
These results demonstrate that the proposed microservice architecture, when properly applied, is well-suited for automotive platforms, offering both performance enhancements and simplified system management.

This article analyzes the performance of containerization, virtualization, and microservice architectures in the context of SDVs, emphasizing their potential for not only improving flexibility but also enhancing overall performance.



{
\bibliographystyle{IEEEtran}
\bibliography{ref}{}
}
\newpage
\begin{IEEEbiography}[{\includegraphics[width=1in,height=1.25in,clip,keepaspectratio]{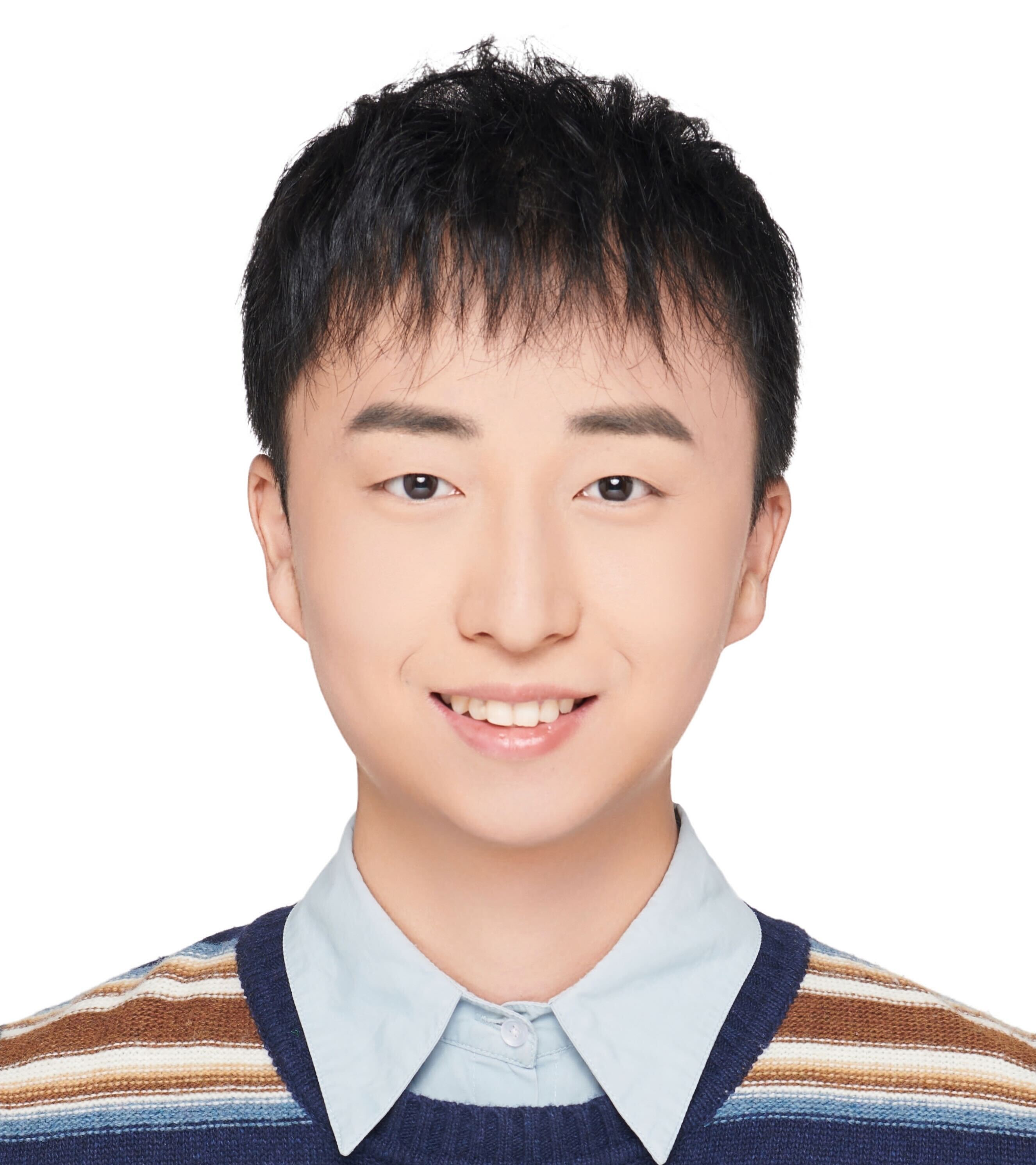}}]{Long Wen}
	received the B.S. degree in 2017 from the school of automotive engineering, Wuhan University of Technology, Wuhan, China. Received the M.S. degree in 2020 from the school of automotive engineering, Jilin University, Changchun, China. He is currently pursuing the Ph.D. with the chair of Robotics, Artificial Intelligence and Real-Time Systems, School of Computation, Information and Technology, Technical University of Munich, Germany.\\ 
	\indent His research interests include autonomous driving, cloud robotic, software engineering, and deep learning.   
\end{IEEEbiography}

\begin{IEEEbiography}[{\includegraphics[width=1in,height=1.25in,clip,keepaspectratio]{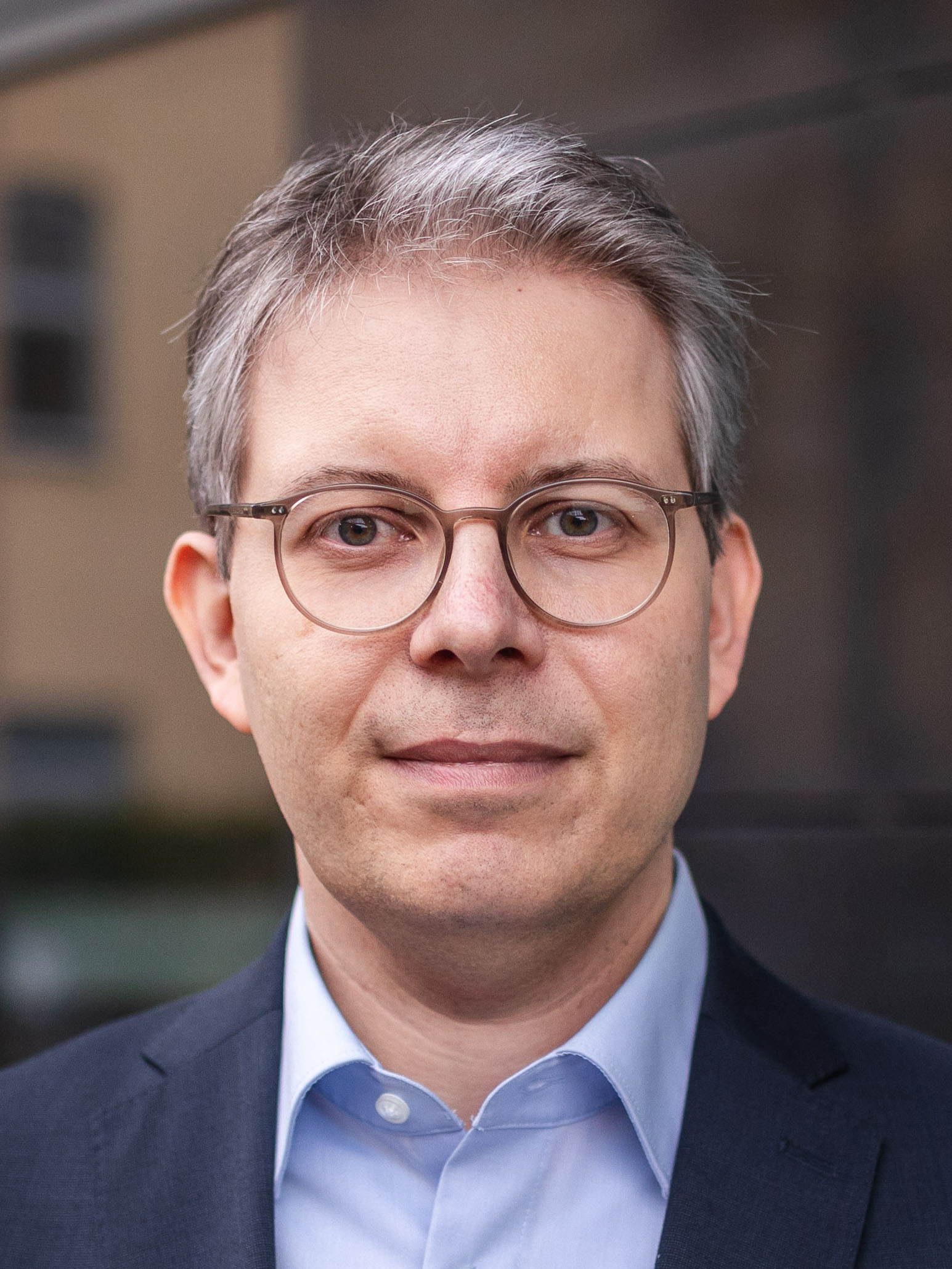}}]{Markus Rickert}
	received his academic degree in Computer Science (Dipl.-Inf. Univ., M.Sc. equiv.) in 2004 and his doctoral degree (Dr. rer. nat., summa cum laude) in 2011, both from the Technical University of Munich (TUM). From 2010 to 2021 he worked at fortiss, affiliated institute of TUM, where he founded the Virtual Engineering and Robotics research group and the Autonomous Driving research group. At fortiss, he was deputy head of Cyber-Physical Systems from 2013 to 2016 and head of Robotics and Machine Learning from 2016 to 2021. From 2021 to 2023 he worked at TUM and was head of the TUM/Huawei Intelligent Automotive Solutions Innovation Lab. Since 2023 he is full professor and chair of Multimodal Intelligent Interaction at the University of Bamberg.\\ 
	\indent His research interests include robotics, human-robot interaction, cognitive systems, artificial intelligence, multimodal interaction, motion planning, advanced systems engineering, and software engineering.	
\end{IEEEbiography}

\begin{IEEEbiography}[{\includegraphics[width=1in,height=1.25in,clip,keepaspectratio]{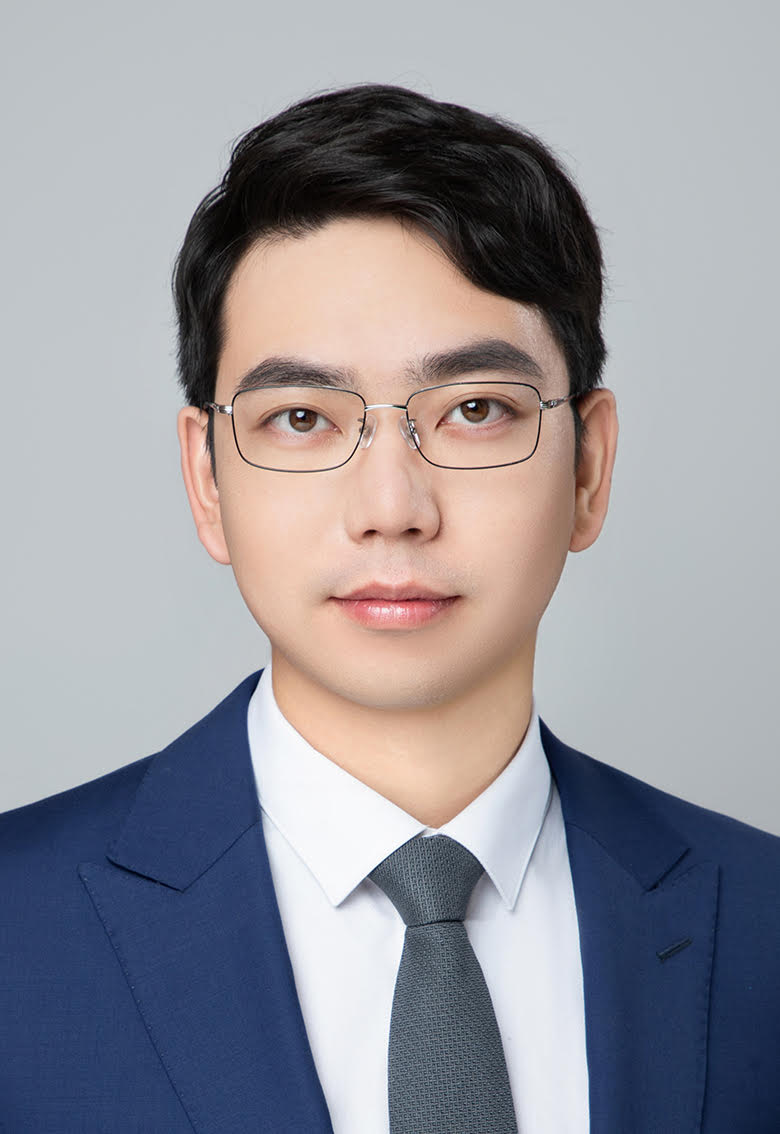}}]{Fengjunjie Pan}
	received the B.Eng. degree in Electrical Engineering from Hamburg University of Applied Sciences in 2016, and the M.Sc. degree in the same major from the Technical University of Berlin in 2019. Afterwards, he worked on continuous integration and testing at Magna Telemotive Munich until 2021. He is currently a research assistant and PhD student at the Chair of Robotics, Artificial Intelligence, and Embedded Systems at Technical University of Munich under the supervision of Prof. Dr.-Ing. habil. Alois Knoll.\\ 
	\indent His research focuses on automotive systems and software engineering, and applying Generative AI to this area.
\end{IEEEbiography}

\begin{IEEEbiography}[{\includegraphics[width=1in,height=1.25in,clip,keepaspectratio]{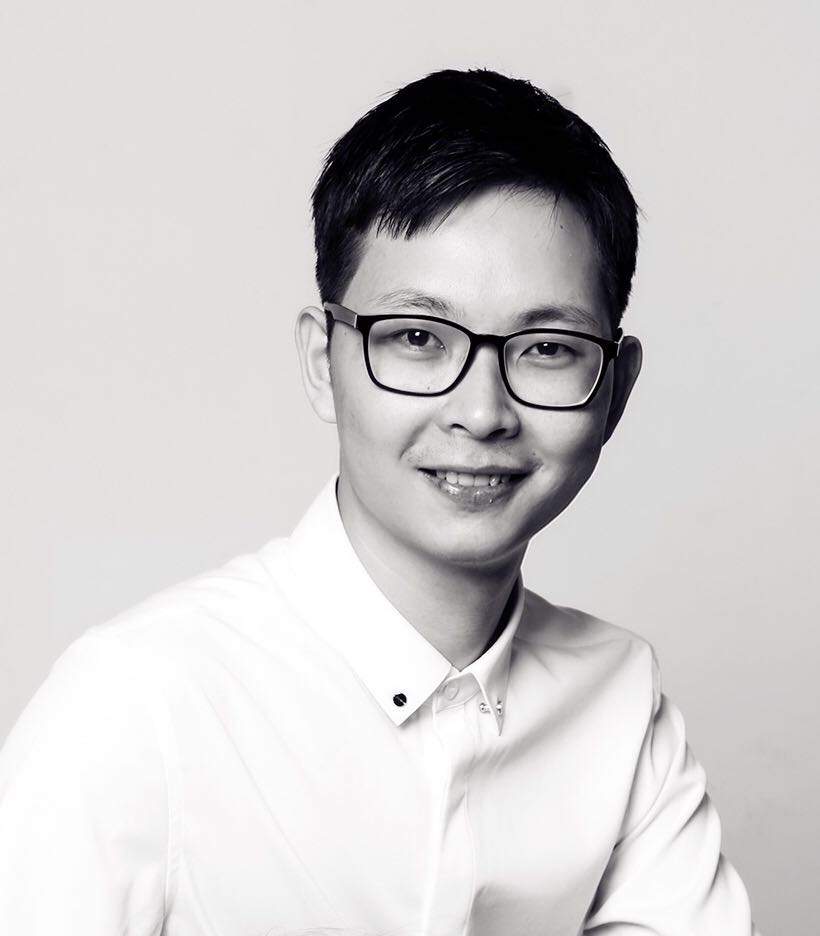}}]{Jianjie Lin}
	Jianjie Lin received a Ph.D. degree in computer science from the Technical University of Munich (TUM) in 2022. He was a Research Associate with the Chair of Robotics, Artificial Intelligence, and Real-Time Systems at TUM, where he has also been a Research Assistant since June 2017. During his studies, he engaged in collaborative research projects with Fortiss GmbH and the German Aerospace Center (DLR), focusing on autonomous systems, motion and trajectory planning, and computer vision.\\ 
	\indent His current research interests include robotics, machine learning, computer vision, software-defined vehicles, autonomous driving, grasp planning, and embodied AI
\end{IEEEbiography}

\begin{IEEEbiography}[{\includegraphics[width=1in,height=1.25in,clip,keepaspectratio]{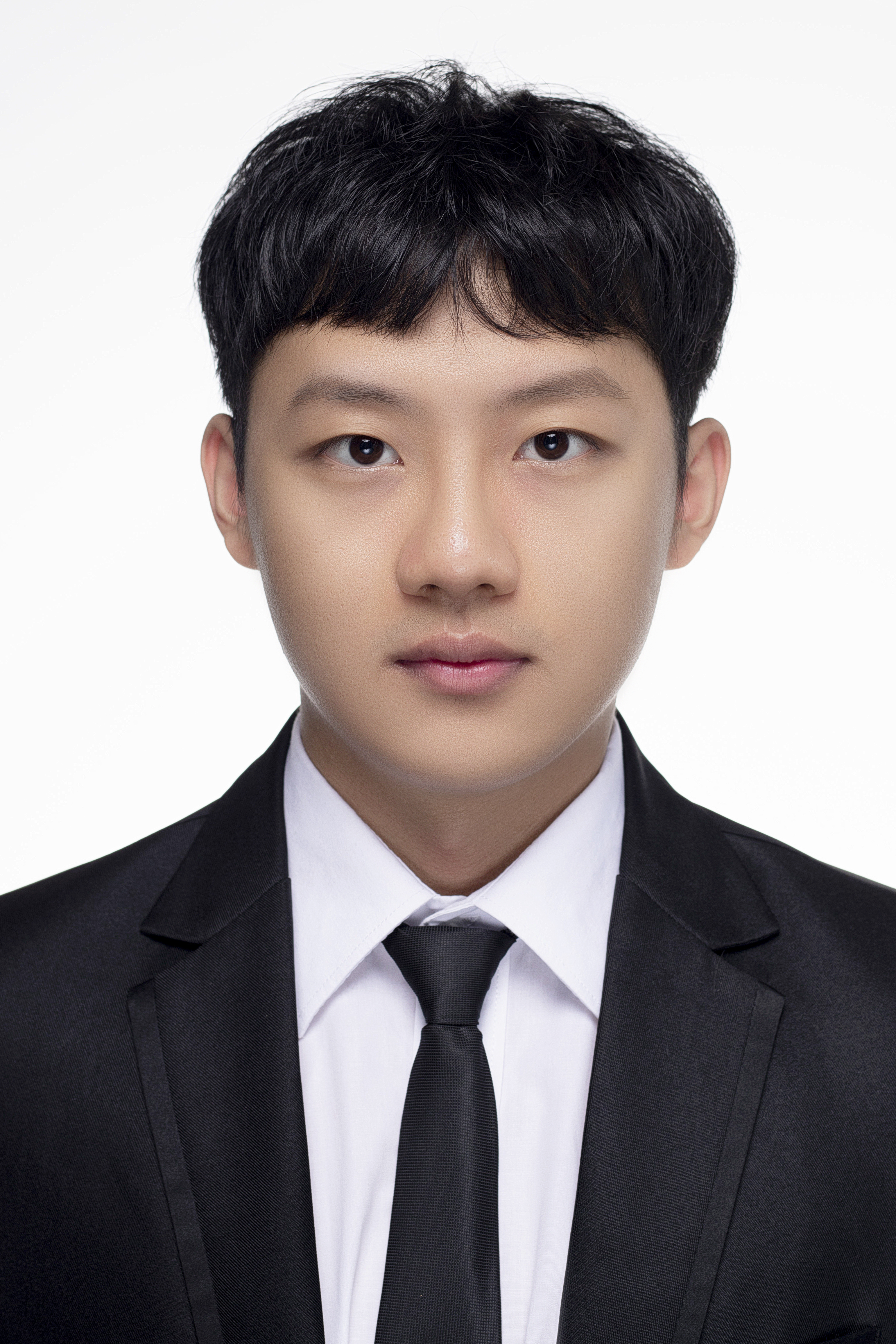}}]{Yu Zhang}%
	received the M.Eng. degree from the School of Intelligence Science and Technology, University of Science and Technology Beijing, Beijing, China, in 2022. He is currently working toward the Ph.D. degree in computer science as a member of the Informatics 6, Technical University of Munich, Munich, Germany.\\
	\indent His current research interests include optimization and control in robotics, machine learning, adaptive and learning control.
\end{IEEEbiography}
\vspace{-105pt}
\begin{IEEEbiography}[{\includegraphics[width=1in,height=1.25in,trim={0 0 0 4.5cm},clip,keepaspectratio]{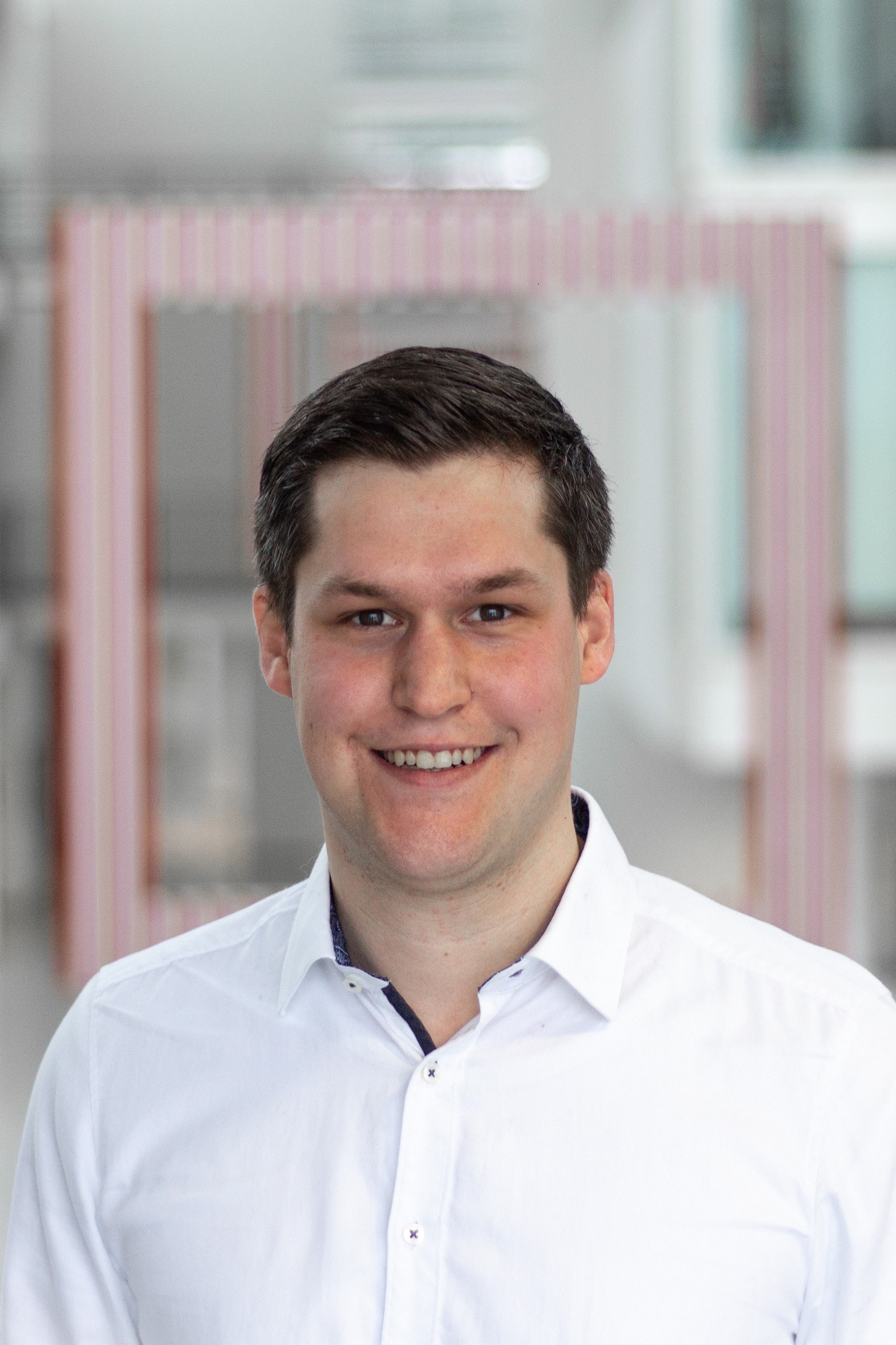}}]{Tobias Betz}
	received the B.Eng. degree from the University of Applied Science Regensburg in 2018, the M.Sc. degree from the Technical University of Munich (TUM) in 2020. He is currently pursuing his Ph.D. at the Institute of Automotive Technology, TUM.\\ 
	\indent His research interests include ROS 2, latency analysis, system optimization, and software deployment on real-world applications in autonomous driving.
\end{IEEEbiography}
\vspace{-105pt}
\begin{IEEEbiography}[{\includegraphics[width=1in,height=1.25in,clip,keepaspectratio]{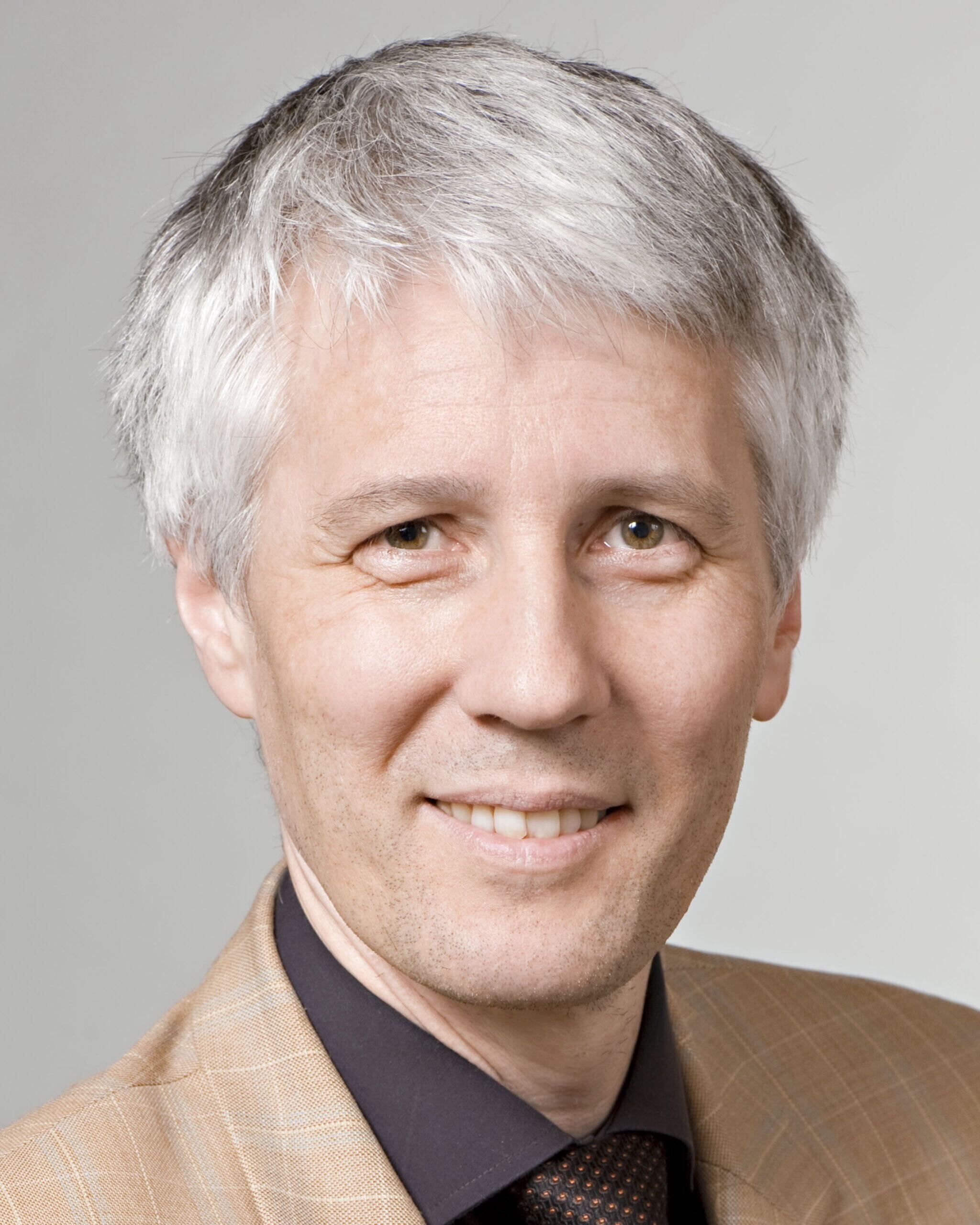}}]{Alois Knoll}
	(IEEE Fellow) received his diploma (M.Sc.) degree in Electrical/Communications Engineering from the University of Stuttgart, Germany, in 1985 and his Ph.D. (summa cum laude) in Computer Science from Technical University of Berlin, Germany, in 1988. He served on the faculty of the Computer Science department at TU Berlin until 1993. He joined the University of Bielefeld, Germany as a full professor and served as the director of the Technical Informatics research group until 2001. Since 2001, he has been a professor at the Department of Informatics, Technical University of Munich (TUM), Germany . He was also on the board of directors of the Central Institute of Medical Technology at TUM (IMETUM). From 2004 to 2006, he was Executive Director of the Institute of Computer Science at TUM. Between 2007 and 2009, he was a member of the EU’s highest advisory board on information technology, ISTAG, the Information Society Technology Advisory Group, and a member of its subgroup on Future and Emerging Technologies (FET). In this capacity, he was actively involved in developing the concept of the EU’s FET Flagship projects.\\ 
	\indent His research interests include cognitive, medical and sensor-based robotics, multi-agent systems, data fusion, adaptive systems, multimedia information retrieval, model-driven development of embedded systems with applications to automotive software and electric transportation, as well as simulation systems for robotics and traffic.
\end{IEEEbiography}

\end{document}